\documentclass[letterpaper]{article} 
\usepackage{aaai2027}  
\usepackage[hyphens]{url}  
\usepackage{graphicx} 
\usepackage{natbib}  
\usepackage{caption} 
\usepackage{cuted}
\usepackage{algorithm}
\usepackage{algorithmic}
\usepackage{amsmath}
\usepackage{amsfonts}
\usepackage{graphicx}
\usepackage{multirow}
\usepackage{makecell}
\usepackage{array}
\usepackage[table]{xcolor}
\usepackage[utf8]{inputenc}
\usepackage{newunicodechar}
\newunicodechar{，}{,}
\usepackage{booktabs}
\usepackage{multirow}
\usepackage{makecell}
\usepackage{amssymb}
\usepackage[table]{xcolor}

\newcommand{\yes}{$\checkmark$}
\newcommand{\no}{\textcolor{gray!45}{$\times$}}
\definecolor{headercream}{RGB}{250,247,239}
\definecolor{sectionblue}{RGB}{232,240,244}
\definecolor{highlightgreen}{RGB}{228,237,211}
\definecolor{spatialbestpink}{RGB}{190,35,100}
\definecolor{spatialsecondpink}{RGB}{225,120,165}

\usepackage[table]{xcolor}
\newcommand{\bestscore}[1]{\textbf{#1}}
\newcommand{\secondscore}[1]{\underline{#1}}
\usepackage{newfloat}
\usepackage{listings}
\DeclareCaptionStyle{ruled}{labelfont=normalfont,labelsep=colon,strut=off} 
\floatstyle{ruled}
\newfloat{listing}{tb}{lst}{}
\floatname{listing}{Listing}

\usepackage{booktabs}
\title{SpatialMind: Adaptive Multi-Level Geometric Fusion for Spatial Reasoning in Multimodal Large Language Models}

\author{
    Shibin Zhao\textsuperscript{\rm 1}\equalcontrib,
    Hao Tan\textsuperscript{\rm 2}\equalcontrib,
    Fan Yang\textsuperscript{\rm 2}\corresponding,
    Haixin Wang\textsuperscript{\rm 2},
    Feilong Chen\textsuperscript{\rm 3},
    Yidong Ma\textsuperscript{\rm 3},
    Chaoyang Zhao\textsuperscript{\rm 2}\corresponding,
    Ming Tang\textsuperscript{\rm 2},
    Jinqiao Wang\textsuperscript{\rm 2}
}
\affiliations{
    \textsuperscript{\rm 1}School of Advanced Interdisciplinary Sciences,
    University of Chinese Academy of Sciences\\
    \textsuperscript{\rm 2}Institute of Automation, Chinese Academy of Sciences\\
    \textsuperscript{\rm 3}Xiaohongshu\\
    Contact: Shibin Zhao (zhaoshibin2025@ia.ac.cn)
}

\usepackage{tabularx}

\newcommand{\vsiresult}[1]{%
    \makebox[\textwidth][c]{%
        \includegraphics[width=0.84\textwidth]{#1}%
    }%
}

\begin{document}
\maketitle
\vspace{-0pt} 
\begin{strip}
    \vspace*{-8pt}
    
    \setlength{\abovecaptionskip}{2pt}
    \includegraphics[width=\textwidth]{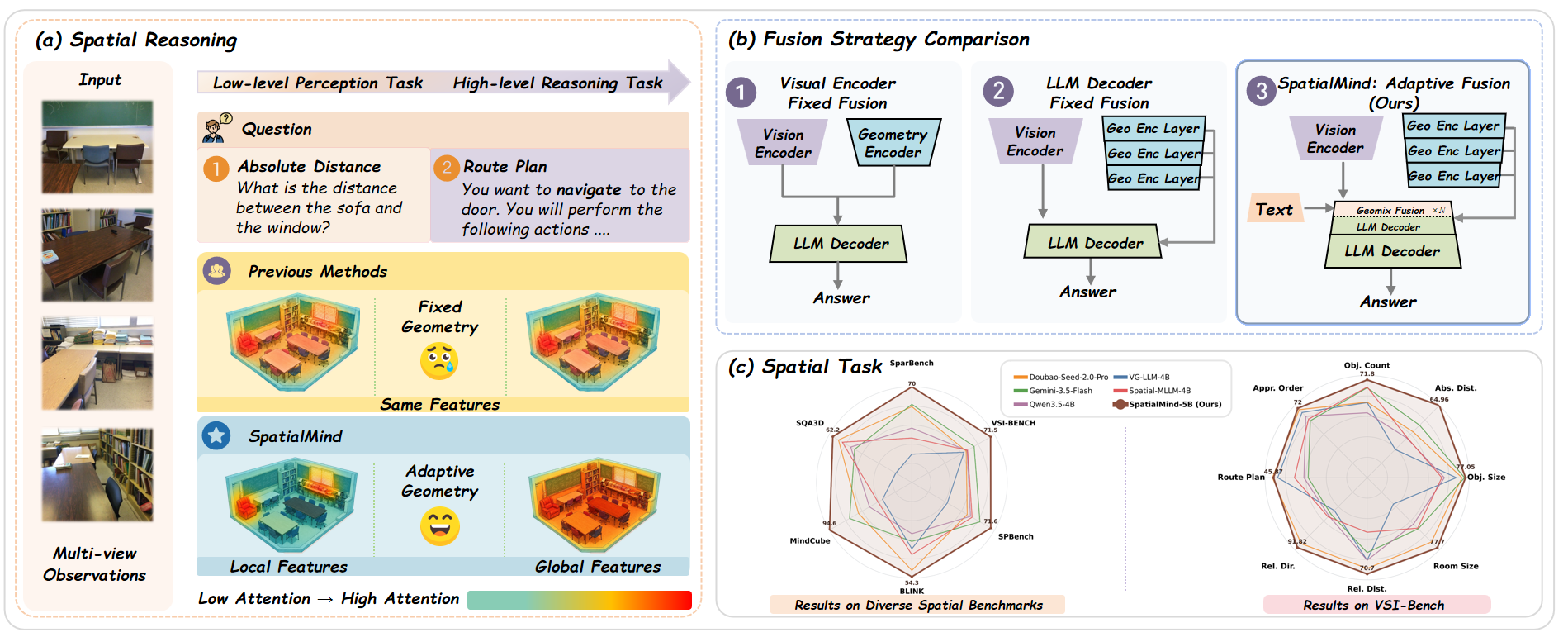}
    \captionof{figure}{{\small \textbf{SpatialMind: Adaptive Multi-Level Geometric Fusion for Spatial Reasoning.} 
    Unlike prior methods that rely on fixed geometric fusion despite the varying granularity requirements of different spatial tasks, SpatialMind adaptively selects task-appropriate features, from local details to global scene structures. (Red and green indicate higher and lower attention values, respectively) \textbf{(b)} Comparison of geometric fusion strategies, with results in \textbf{(c)} showing that SpatialMind achieves competitive performance across spatial intelligence benchmarks.}}

    \label{fig:main}
\end{strip}
\begin{abstract}
Multimodal Large Language Models (MLLMs) demonstrate promising capabilities in 2D visual understanding, yet remain limited in 3D spatial reasoning. Recent studies explore integrating 3D reconstruction models into MLLMs. However, these approaches either use a single geometric granularity across all tasks or adopt a fixed multi-level fusion scheme, regardless of task-specific reasoning requirements. Through systematic experiments on multi-level geometric fusion, we find that spatial tasks at different cognitive levels (e.g., distance estimation vs. route planning) require geometric representations of different granularities, revealing that fixed fusion cannot accommodate diverse spatial reasoning requirements. To bridge this gap, we propose \textit{SpatialMind}, a task-adaptive framework for dynamically integrating multi-level geometric representations. SpatialMind employs GeoFus, a multi-layer geometric fusion module built around a gating network jointly conditioned on textual instructions, visual semantics, and geometric representations. This network adaptively modulates the contributions of shallow, intermediate, and deep geometric features across the layers of the LLM decoder. Furthermore, we introduce a negative-entropy regularization loss to mitigate gate collapse. Gate visualizations indicate that GeoFus adapts geometric granularity to task requirements. Experiments demonstrate that \textit{SpatialMind} achieves competitive performance across spatial benchmarks, attaining a state-of-the-art score of 72.3 on VSI-Bench.
The source code is publicly available at
\url{https://github.com/zhaosb-star/SpatialMind}.
\end{abstract}
\section{Introduction}
\indent
In recent years, Multimodal Large Language Models (MLLMs) demonstrate remarkable capabilities in open-world visual understanding and reasoning \cite{alayrac2022flamingo,li2023blip,liu2023visual,bai2023qwenvl}. With the rapid development of embodied and spatial intelligence, MLLMs increasingly need capabilities beyond general 2D visual semantics. These capabilities, ranging from the perception of object locations, geometric distances, and topological structures to navigation and interaction, are fundamental to embodied intelligence. However, existing MLLM architectures are primarily designed for 2D visual-semantic alignment, which constrains their ability to perceive and reason about 3D spatial structures (Yang et al. 2025a; Stogiannidis, McDonagh, and Tsaftaris 2025; Fu et al. 2024). Therefore, spatial intelligence requires the effective integration of 3D geometric priors into vision-language representations without compromising the semantic capabilities of MLLMs. This challenge has motivated growing interest in more effective multimodal fusion paradigms.
\par
\indent
To enhance the 3D spatial understanding of MLLMs, some approaches incorporate auxiliary geometric modalities as additional inputs \cite{hong20233d,deng20253d,zhu2024llava}. As illustrated in Fig.~\ref{fig:main}(b), other methods employ geometry encoders and fuse their representations with visual features or inject them into multiple layers of the LLM decoder \cite{zheng2026learning,fan2026vlm,wu2026spatial,zhao2025spacemind,zhang2026spatialstack,li2026thinking}. However, these methods either rely on a single level of geometric representation or adopt a fixed multi-level fusion scheme across all spatial tasks. Such task-agnostic strategies overlook the varying requirements for geometric representations of different spatial tasks and cannot fully exploit multi-level geometric cues. This raises a critical question: how can a fusion architecture adaptively match the granularity and integration scheme required by each spatial task?
\par
\indent
We conduct systematic experiments to answer this question, injecting multi-level geometric features into different layers of the LLM decoder. For evaluation, we follow the spatial cognitive hierarchy defined in SPAR \cite{zhang2026flatland} and divide VSI-Bench \cite{yang2025thinking} into low-level perception and high-level reasoning tasks. As shown in Fig.~\ref{fig:Effects of Multi-Level Geometric Features.}, shallow features better support low-level tasks, while deeper features better support high-level tasks. However, previous fixed fusion methods cannot adaptively adjust the relative contributions of these features based on task requirements, as illustrated in Fig.~\ref{fig:main}(a). This task-dependent behavior is consistent with findings in cognitive neuroscience, where the brain dynamically reallocates activation between low- and high-level neural pathways according to task complexity \cite{goodale1992separate}. Therefore, we argue that each LLM injection layer should adaptively aggregate multi-level geometric representations to match the spatial granularity required by the task.
\par
\indent
We propose SpatialMind, a task-adaptive framework for spatial reasoning that employs GeoFus, a multi-layer geometric feature fusion module. GeoFus employs a gating network that jointly models textual instructions, visual semantics, and spatial information. This network dynamically modulates the contributions of shallow, intermediate, and deep geometric features across different LLM injection layers, enabling the model to better adapt feature granularity to task-specific requirements. To mitigate gate collapse, we further design a negative-entropy regularization loss. Visualizations of the learned gating weights show that GeoFus selects geometric representations at task-appropriate granularities. SpatialMind achieves competitive results across spatial reasoning benchmarks, with its 5B and 10B variants setting state-of-the-art scores of 71.5 and 72.3 in their respective model-scale groups on VSI-Bench. The main contributions of this work are summarized as follows:
\par

\begin{itemize}
    \item We analyze the limitations of fixed geometry-to-reasoning pairings and reveal that spatial tasks at different cognitive levels require geometric representations of varying granularities and integration patterns.
    
    \item We propose SpatialMind, a task-adaptive framework for spatial reasoning with GeoFus, which modulates the contributions of multi-level geometric representations across LLM injection layers and adopts a negative-entropy regularization objective to mitigate gate collapse.
    
    \item \textit{SpatialMind} achieves competitive performance across multiple spatial reasoning benchmarks and establishes a new state of the art on VSI-Bench with a score of 72.3.
\end{itemize}

\section{Related Work}
\subsection{Multimodal Large Language Models}
MLLMs extend the capabilities of LLMs to visual understanding by aligning visual and textual representations in a shared multimodal space. Early frameworks like CLIP \cite{radford2021learning} and ALIGN \cite{jia2021scaling} employed contrastive learning to support zero-shot transfer. Subsequently, paradigms such as Flamingo \cite{alayrac2022flamingo} and BLIP-2 \cite{li2023blip} introduced lightweight interfaces, utilizing Q-Formers or cross-attention mechanisms to connect pre-trained visual encoders with LLMs. Driven by the advancement of visual instruction tuning, models like InstructBLIP \cite{dai2023instructblip}, LLaVA \cite{liu2023visual,liu2024improved}, and Qwen-VL \cite{bai2023qwenvl} employed projection layers to bridge pre-trained visual encoders and LLMs, while exploring more efficient representation and fusion paradigms. Recently, video-based MLLMs \cite{lin2024video,zhang2024llavavideo,li2024llava} have extended this paradigm to the temporal domain through spatiotemporal token compression, temporal attention, and long-context modeling, achieving substantial progress in general video understanding. Despite these advances, existing MLLMs remain limited in spatial reasoning. Most rely on final-layer semantic features and frame-wise 2D tokens, overlooking fine-grained geometry and explicit modeling of depth, camera motion, and cross-view consistency \cite{meng2024deepstack,oquab2023dinov2}. Consequently, existing MLLMs lack robust 3D spatial representations, limiting their ability to generalize across diverse geometric perception and spatial reasoning tasks.

\subsection{MLLMs for Spatial Intelligence}
To enhance the spatial intelligence of MLLMs, prior studies have explored various ways of incorporating geometric information \cite{schonberger2016structure,yao2018mvsnet,ranftl2021vision,wang2024dust3r,leroy2024grounding,wang2025continuous,wang2025vggt}. Early approaches incorporated auxiliary modalities to enhance 3D spatial understanding. For instance, LLaVA-3D \cite{zhu2024llava} and Video-3D LLM \cite{zheng2025video} leverage lightweight RGB-D encoders to capture explicit geometric cues. However, their dependence on depth sensors or pre-scanned environments limits their applicability to standard RGB videos. To alleviate this dependence, recent methods integrate geometric representations directly into MLLM architectures. VG-LLM \cite{zheng2026learning} augments visual tokens with patch-level geometric features, whereas SpatialStack \cite{zhang2026spatialstack} injects multi-level geometric representations into selected LLM decoder layers. Nevertheless, these methods typically rely on fixed and task-agnostic fusion schemes, which may introduce conflicts between semantic and geometric representations. Another line of research transfers geometric knowledge through feature distillation or representation alignment. Frameworks like 3DRS \cite{huang20263drs} and 3DThinker \cite{chen2026think} distill the spatial priors of 3D representation models into the visual representations of MLLMs. Although such alignment improves geometric awareness, it does not explicitly support sample-dependent selection of geometric information during inference. Overall, existing methods apply the same predefined geometric feature levels and injection pathways across tasks. To resolve these limitations, this paper proposes an adaptive feature fusion strategy to achieve task-adaptive integration of multi-level geometric features.

\begin{figure}[t]
    
    \captionsetup{skip=0pt}
    \includegraphics[width=\columnwidth]{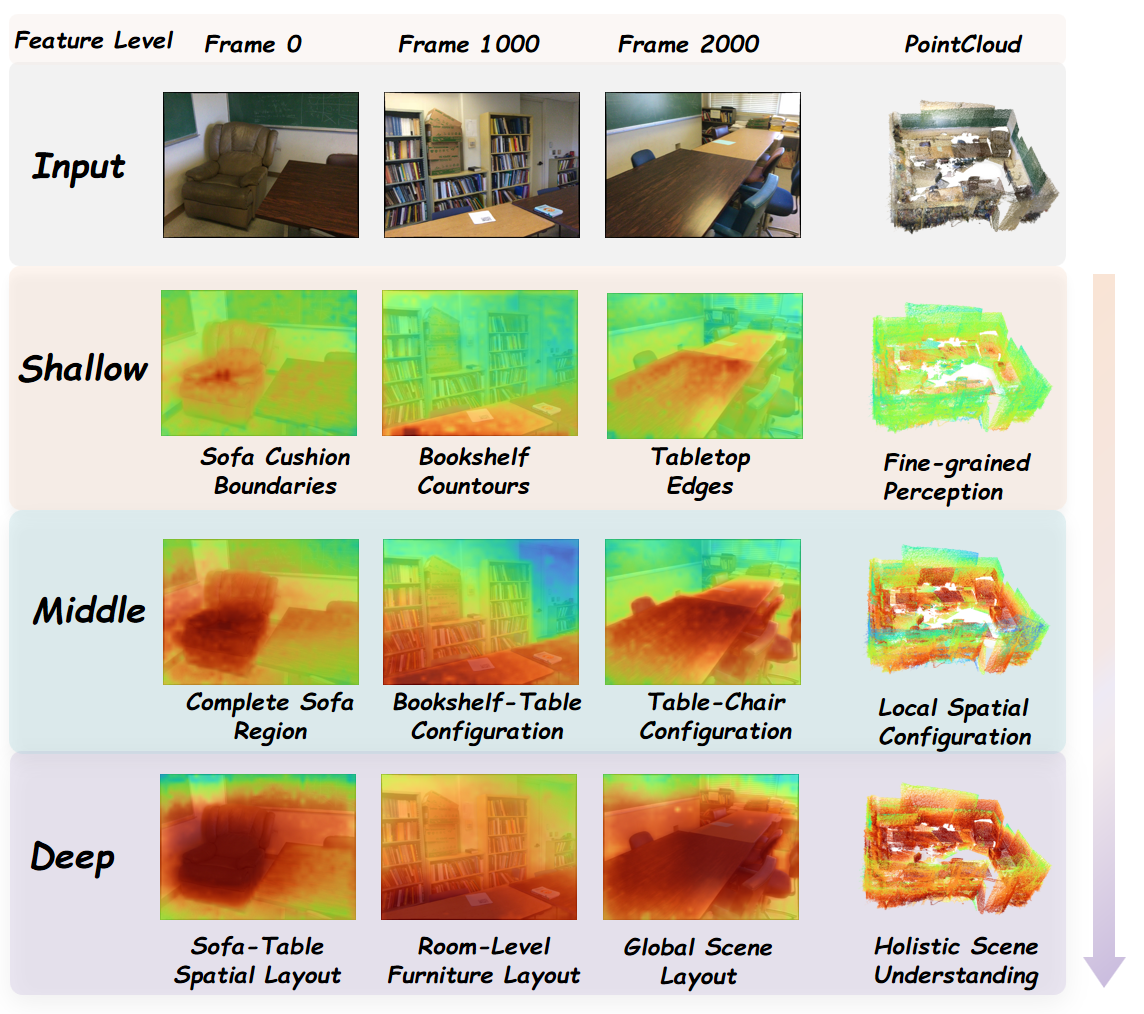}
    \caption{\small
    \textbf{Visualization of Multi-Level Geometric Attention across 2D Views and 3D Point Clouds.} Features at different levels capture spatial information at distinct scales, progressing from fine-grained details to global layouts.
    }
    \label{fig:Attention Visualization.}
\end{figure}

\section{Hierarchical Roles of Multi-Level Geometric Features}
To investigate how the geometric granularity required by spatial tasks varies with cognitive complexity, we conduct controlled ablation experiments on VSI-Bench \cite{yang2025thinking} using \textit{SpatialStack} as the baseline. With the model architecture, training settings, and injection routes unchanged, we separately remove the shallow injection (VGGT-11 to LLM-0), intermediate injection (VGGT-17 to LLM-1), and deep injection (VGGT-23 to LLM-2), and evaluate the performance of these configurations across various spatial tasks. Following the spatial cognitive difficulty spectrum defined in SPAR \cite{zhang2026flatland}, we divide VSI-Bench into \textit{low-level tasks}, including relative distance, absolute distance, relative direction, object size, and room size, and \textit{high-level tasks}, including route planning, object appearance order, and object counting.
This analysis examines two hypotheses: \textbf{(i) Cross-level Complementarity}, whether the complete multi-level configuration outperforms variants with any geometric layer removed, indicating that features from different levels provide complementary information; and \textbf{(ii) Relative Functional Specialization}, whether features at different levels contribute differently to distinct task hierarchies, suggesting that multi-level geometric representations serve specialized functional roles.

\begin{figure}[t]
    
    \captionsetup{skip=0pt} 
    \includegraphics[width=\columnwidth]{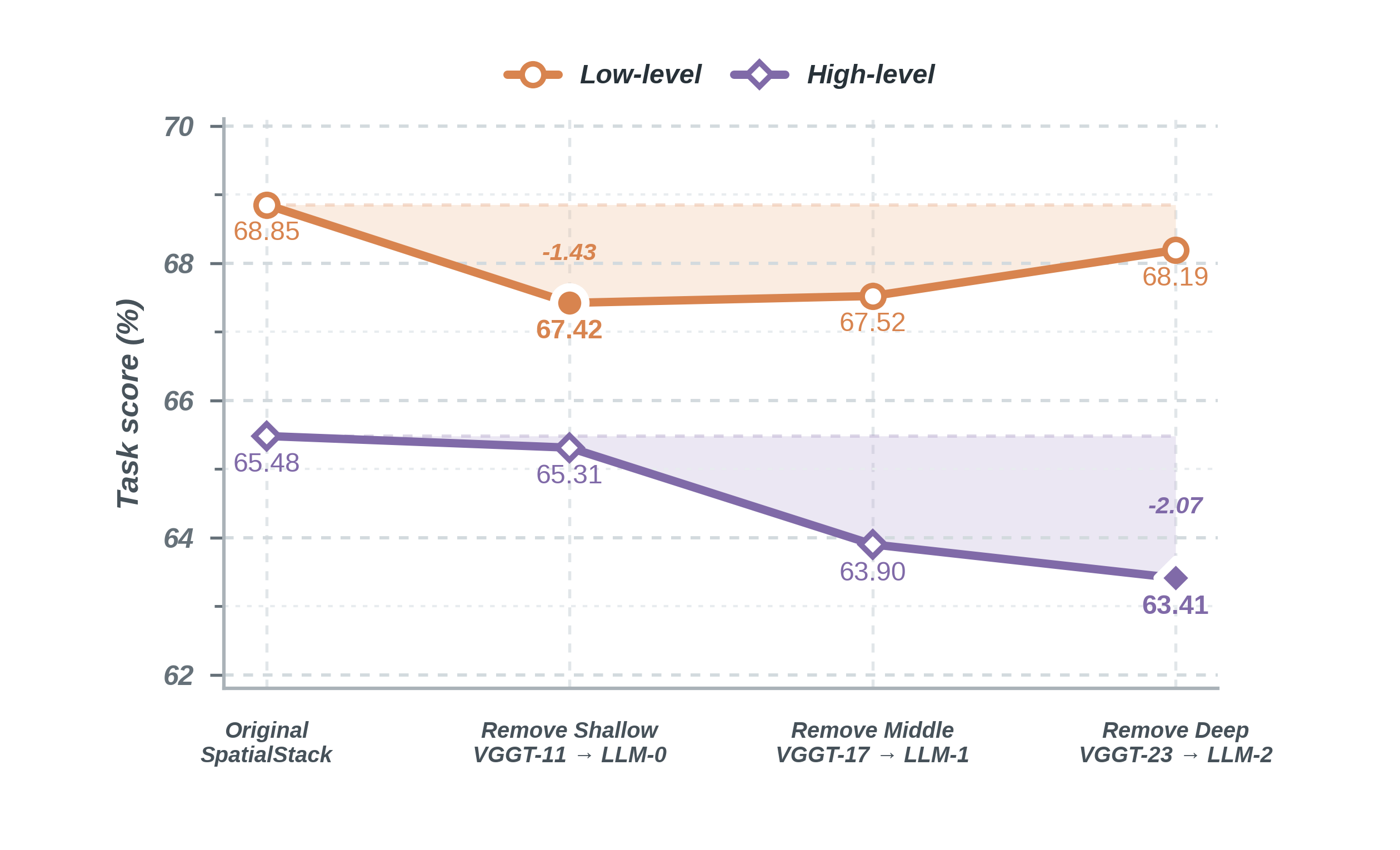} 
    \caption{\small
    \textbf{Effects of Multi-Level Geometric Features.}
    Shallow features contribute more to low-level tasks, while deep features are more important for high-level tasks, demonstrating hierarchical specialization and cross-level complementarity.
    }
    \label{fig:Effects of Multi-Level Geometric Features.}
\end{figure}

\textbf{Qualitative Analysis.} To examine the hierarchical differences among multi-level geometric features, we visualize their attention patterns on scene0030\_00 from ScanNet  \cite{dai2017scannet}. As shown in Fig.~\ref{fig:Attention Visualization.}, shallow features respond primarily to object boundaries, local contours, and depth discontinuities, preserving fine-grained geometric details. Meanwhile, intermediate features exhibit an expanded focus that covers more holistic object regions, beginning to consolidate object shapes and local spatial configurations. Ultimately, deep features exhibit more globally distributed attention patterns, emphasizing inter-object relationships and the overall scene layout. These observations indicate that multi-level geometric features form a progressive representation hierarchy, evolving from fine-grained perception to local spatial configurations and ultimately to global spatial understanding.
\par

\textbf{Quantitative Results.} The quantitative results presented in Fig.~\ref{fig:Effects of Multi-Level Geometric Features.} further corroborate the aforementioned qualitative insights. The complete multi-level configuration achieves the best performance on both low- and high-level tasks, demonstrating that features at different levels provide highly complementary information. Removing the shallow injection reduces low-level task performance by $1.43$ percentage points, while causing only a $0.17$ point drop on high-level tasks. This suggests that shallow features contribute more strongly to fundamental spatial attributes, such as distance, direction, and scale. In contrast, removing the deep injection decreases high-level task performance by $2.07$ percentage points, compared with a smaller drop of $0.66$ points on low-level tasks. This result indicates that deep features play a more important role in modeling global scene context and higher-level spatial relationships. Removing the intermediate injection leads to comparable drops of $1.33$ and $1.58$ percentage points on low- and high-level tasks, respectively. These results suggest that intermediate features provide stable support across both task hierarchies, bridging local geometric perception, object-level relations, and global spatial reasoning.
\par

\begin{figure*}[t]
    
    \includegraphics[
        width=0.98\textwidth,
        keepaspectratio
    ]{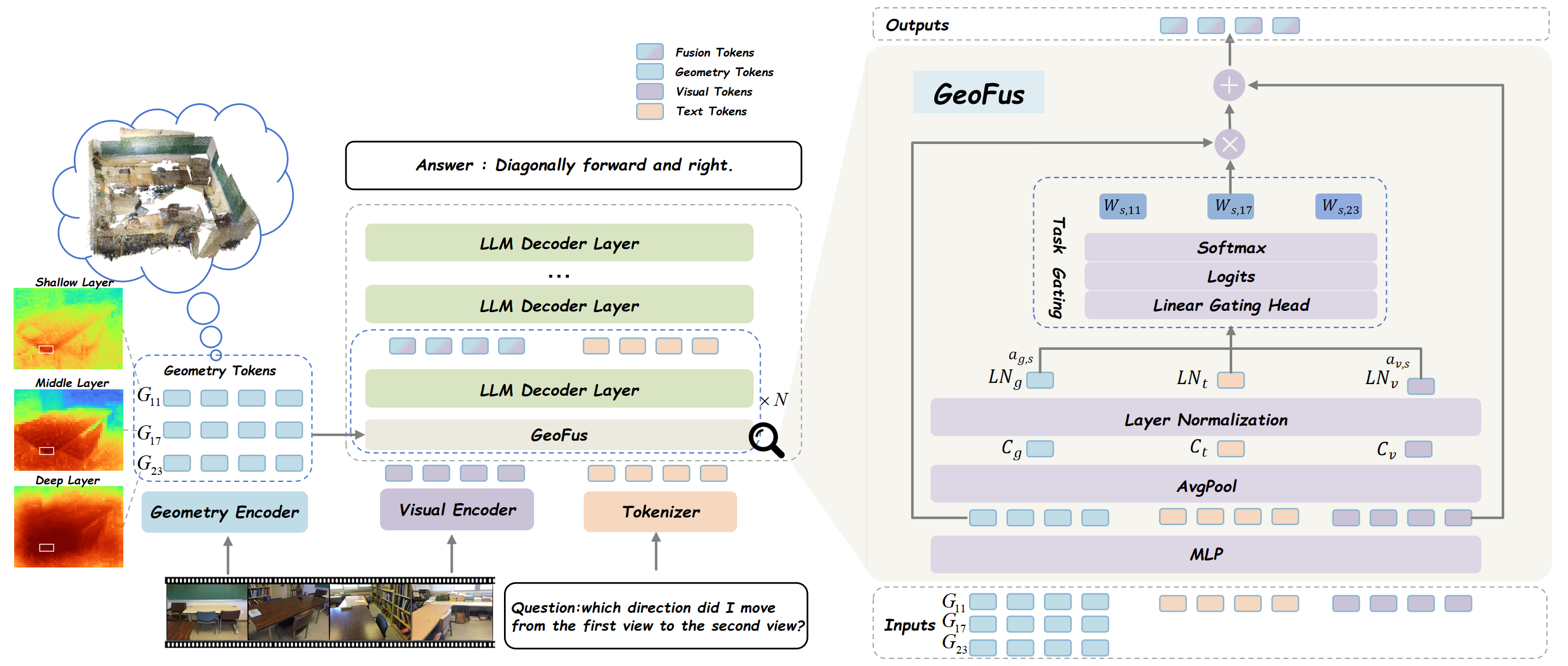}
    \caption{{\small \textbf{Overview of the \textit{SpatialMind} architecture.} 
   Our framework adopts a decoupled fusion design and introduces \textbf{GeoFus} for task-adaptive integration of multi-level geometric features with visual tokens. By jointly conditioning on textual, visual, and geometric contexts, GeoFus dynamically adjusts the contributions of geometric representations at different LLM injection layers according to task requirements.}}
    \label{fig:architecture}
\end{figure*}

The visualization and quantitative results reveal hierarchical specialization and cross-level complementarity, motivating each LLM injection layer to adaptively fuse the geometric granularity most relevant to the current spatial task.

\section{Method}
We propose \textit{SpatialMind}, a task-adaptive framework for spatial reasoning that dynamically modulates the contributions of multi-level geometric representations according to task-specific spatial reasoning requirements.
Sec.~4.1 introduces the overall architecture, Sec.~4.2 describes GeoFus, and Sec.~4.3 presents the negative-entropy regularization loss.
\par

\begin{table*}[t]
    
    \caption{\small{\textbf{Evaluation on VSI-Bench.}
    \bestscore{Bold} and \secondscore{underlined} values indicate the best
    and second-best results within each open-source spatial intelligence
    model scale group, respectively.
    \textsuperscript{\(\dagger\)} denotes models trained using the same
    training data.}
    }
    \label{tab:vsibench_main_results}

    \begingroup
    \renewcommand{\arraystretch}{0.95}
    \setlength{\tabcolsep}{2.8pt}
    \setlength{\aboverulesep}{1.4pt}
    \setlength{\belowrulesep}{1.4pt}
    \fontsize{5.3}{6.5}\selectfont
    \setlength{\abovecaptionskip}{1pt}

    \resizebox{\textwidth}{!}{%
    \begin{tabular}{
        >{\raggedright\arraybackslash}p{4.8cm}
        c
        c
        *{8}{c}
    }

        \toprule

        \multirow{2}{*}{\textbf{Methods}}
        & \multirow{2}{*}{\textbf{Rank}}
        & \multirow{2}{*}{\textbf{Avg.}}
        & \multicolumn{4}{c}{
            \cellcolor{headercream}\textbf{Numerical Answer}
        }
        & \multicolumn{4}{c}{
            \cellcolor{headercream}\textbf{Multiple-Choice Answer}
        }
        \\

        \cmidrule(lr){4-7}
        \cmidrule(lr){8-11}

        & &
        & \makecell{\textbf{Obj.}\\\textbf{Count}}
        & \textbf{Abs. Dist.}
        & \textbf{Obj. Size}
        & \makecell{\textbf{Room}\\\textbf{Size}}
        & \textbf{Rel. Dist.}
        & \textbf{Rel. Dir.}
        & \makecell{\textbf{Route}\\\textbf{Plan}}
        & \makecell{\textbf{Appr.}\\\textbf{Order}}
        \\

        \rowcolor{sectionblue}
        \multicolumn{11}{l}{
            \textbf{\textit{Proprietary Models (API)}}
        }
        \\

        Doubao-Seed-2.0-Pro~\cite{seed2026seed2}
        & 3 & 49.3
        & 55.1 & 25.3 & 69.8 & 29.1
        & 60.6 & 43.2 & 43.8 & 67.6
        \\

        GPT-5.6-Luna~\cite{openai_gpt56_2026}
        & 2 & 52.2
        & 49.6 & 26.2 & 71.2 & 53.2
        & 56.6 & 47.2 & 47.4 & 66.2
        \\

        Gemini-3.5-Flash~\cite{gemini35flash_2026}
        & 1 & 56.5
        & 55.5 & 41.3 & 74.3 & 58.0
        & 66.6 & 50.3 & 38.1 & 67.8
        
        \\

        \midrule

        \rowcolor{sectionblue}
        \multicolumn{11}{l}{
            \textbf{\textit{Open-source Models}}
        }
        \\

        InternVL3-78B~\cite{zhu2025internvl3}
        &  & 48.5
        & 71.2 & 53.7 & 44.4 & 39.5
        & 55.9 & 39.5 & 28.9 & 54.5
        \\

        LLaVA-Video-7B~\cite{zhang2024llavavideo}
        &  & 35.6
        & 48.5 & 14.0 & 47.8 & 24.2
        & 43.5 & 42.4 & 34.0 & 30.6
        \\

        LLaVA-Video-72B
        &  & 40.9
        & 48.9 & 22.8 & 57.4 & 35.3
        & 42.4 & 36.7 & 35.0 & 48.6
        \\

        Qwen2.5-VL-7B
        &  & 31.5
        & 31.2 & 9.2 & 39.9 & 40.9
        & 38.7 & 38.9 & 28.4 & 24.8
        \\

        Qwen3.5-4B~\cite{qwen35blog}
        &  & 50.1
        & 47.6 & 35.6 & 60.5 & 51.4
        & 60.6 & 50.3 & 30.9 & 63.8
        \\

        Qwen3-VL-4B~\cite{bai2025qwen3}
        & 3 & 54.6
        & 66.0 & 47.2 & 76.3 & 56.6
        & 54.9 & 47.7 & 28.9 & 59.1
        \\

        Qwen3.5-9B
        & 2 & 55.7
        & 51.0 & 36.7 & 68.9 & 63.0
        & 63.9 & 55.2 & 39.2 & 67.5
        \\

        Qwen3.5-4B\textsuperscript{\(\dagger\)}
        & 1 & 66.5
        & 71.1
        & 52.6
        & 77.2
        & 72.5
        & 64.2
        & 79.7
        & 38.7
        & 76.2
        \\
        
        \midrule

        \rowcolor{sectionblue}
        \multicolumn{11}{l}{
            \textbf{\textit{
                Specialized Spatial Reasoning Models (3--5B)
            }}
        }
        \\

        Spatial-MLLM-4B~\cite{wu2026spatial}
        &  & 50.6
        & 66.5 & 36.0 & 58.5 & 55.9
        & 40.0 & 51.4 & 35.6 & 60.8
        \\

        VG-LLM-4B~\cite{zheng2026learning}
        &  & 47.3
        & 66.0 & 37.8 & 55.2 & 59.2
        & 44.6 & 45.6 & 33.5 & 36.4
        \\

        Cambrian-S-3B~\cite{yang2025cambrian}
        &  & 57.3
        & 70.7 & 40.6 & 68.0 & 46.3
        & 64.8 & 61.9 & 27.3 & \secondscore{78.8}
        \\

        SpatialStack-5B~\cite{zhang2026spatialstack}
        & 3 & 67.1
        & 70.2
        & 55.5
        & 70.9
        & 67.4
        & \secondscore{66.9}
        & 82.1
        & 41.8
        & \bestscore{81.7}
        \\

        SpatialStack-5B\textsuperscript{\(\dagger\)}
        & 2
        & \secondscore{69.6}
        & \secondscore{70.7}
        & \secondscore{60.6}
        & \secondscore{76.7}
        & \secondscore{73.1}
        & 66.3
        & \secondscore{88.9}
        & \bestscore{46.9}
        & 73.3
        \\

        \textbf{SpatialMind-5B (ours)}
        & 1
        & \bestscore{71.5}
        & \bestscore{71.8}
        & \bestscore{65.0}
        & \bestscore{77.1}
        & \bestscore{77.7}
        & \bestscore{70.7}
        & \bestscore{91.8}
        & \secondscore{45.9}
        & 72.0
        \\

        \midrule

        \rowcolor{sectionblue}
        \multicolumn{11}{l}{
            \textbf{\textit{
                Specialized Spatial Reasoning Models (8--10B)
            }}
        }
        \\

        GeoThinker-8B (Qwen2.5-VL-7B)~\cite{li2026thinking}
        &  & 67.9
        & 72.4 & 55.1 & 74.6 & 71.2
        & 65.9 & 86.7 & 46.4 & 70.6
        \\

        SpatialStack-10B
        &  & 68.8
        & 70.8
        & 55.5
        & 71.9
        & 70.1
        & 68.7
        & 85.3
        & \secondscore{47.9}
        & \bestscore{80.4}
        \\

        SpaceMind-8B~\cite{zhao2025spacemind}
        &  & 69.6
        & \secondscore{73.3}
        & 61.4
        & \secondscore{77.3}
        & 74.2
        & 67.2
        & 88.4
        & 44.3
        & 70.6
        \\

        SpatialStack-10B\textsuperscript{\(\dagger\)}
        & 3
        & 71.3
        & 72.1
        & \secondscore{62.3}
        & \bestscore{78.1}
        & \secondscore{74.9}
        & 70.7
        & 90.0
        & 42.3
        & \secondscore{79.8}
        \\

        GeoThinker-9B (Qwen3-VL-8B)
        & 2
        & \secondscore{71.9}
        & \bestscore{73.7}
        & 61.5
        & 76.2
        & 73.8
        & \bestscore{75.1}
        & \secondscore{90.0}
        & \bestscore{49.5}
        & 75.4
        \\

        \textbf{SpatialMind-10B (ours)}
        & 1
        & \bestscore{72.3}
        & 72.4
        & \bestscore{66.1}
        & 77.0
        & \bestscore{77.4}
        & \secondscore{72.8}
        & \bestscore{92.5}
        & 46.9
        & 73.6
        \\

        \bottomrule

    \end{tabular}%
    }

    \endgroup
\end{table*}

\subsection{VLM-SpatialMind}
\subsubsection{VLM Architecture.}
Given a sequence of $K$ images $\mathcal{I} = \{I_k \in \mathbb{R}^{H \times W \times 3}\}_{k=1}^K$ and a text query $Q$, we extract frame-wise visual representations via a shared visual encoder: $\mathbf{V}_k = \text{Encoder}_{2\text{D}}(I_k) \in \mathbb{R}^{N_{\text{vis}} \times D_{\text{vis}}}$, where $N_{\mathrm{vis}}=HW/p^2$ is the number of visual tokens per frame
for a patch size of $p$, and $D_{\text{vis}}$ is the hidden dimension of the 2D visual encoder. To handle high-resolution inputs efficiently, a spatial merger aggregates $2 \times 2$ neighboring visual tokens with stride $2$ into downsampled features: $\tilde{\mathbf{V}}_k = \text{SpatialMerger}(\mathbf{V}_k) \in \mathbb{R}^{N' \times D_{\text{lang}}}$, where $N'=HW/(4p^2)$ and $D_{\mathrm{lang}}$ is the hidden dimension of the LLM. We then concatenate all frame tokens and text embeddings $\mathbf{T} \in \mathbb{R}^{M \times D_{\text{lang}}}$ to construct the initial input $\mathbf{H}_0 = [ \tilde{\mathbf{V}}_1; \dots; \tilde{\mathbf{V}}_K ; \mathbf{T} ] \in \mathbb{R}^{(KN' + M) \times D_{\text{lang}}}$, which is subsequently processed by an $L$-layer LLM decoder.

\subsubsection{3D Visual Geometry Encoder.} To acquire spatial priors without requiring additional 3D annotations during SpatialMind training, we employ a frozen pretrained VGGT \cite{wang2025vggt} as the geometry encoder. For each image $I_k$, VGGT introduces camera token $\mathbf{c}_k$ and register tokens $\mathbf{r}_k$ alongside patch spatial tokens $\mathbf{p}_k$, forming the sequence $\mathbf{Z}_0^{(k)} = [\mathbf{c}_k; \mathbf{r}_k; \mathbf{p}_k] \in \mathbb{R}^{(1+R+N_{\text{geo}}) \times D_{\text{geo}}}$, where $N_{\mathrm{geo}}$ denotes the number of geometric patch tokens and $D_{\mathrm{geo}}$ is the hidden dimension of VGGT. Feeding these concatenated sequences into $L_{\text{geo}}$ Transformer layers extracts geometry-enriched features. For each selected layer $\ell$, we discard the camera token and register tokens and retain only the patch-token features $\mathbf{Z}_{\ell}\in\mathbb{R}^{(KN_{\mathrm{geo}})\times D_{\mathrm{geo}}}$. To align these features with the visual-token resolution and hidden
dimension of the multimodal backbone, we apply a layer-specific geometric merger with a spatial stride of $2$: $\mathbf{G}_{\ell}
=\mathcal{M}_{\mathrm{geo}}^{(\ell)}(\mathbf{Z}_{\ell})
\in\mathbb{R}^{(KN')\times D_{\mathrm{lang}}},\qquad\ell\in\mathcal{K}=\{11,17,23\}$. These selected layers correspond to shallow, intermediate, and deep VGGT representations, respectively, forming a progressive geometric hierarchy from fine-grained details to object-level configurations and global scene layouts.

\subsection{GeoFus} 
We propose \textbf{GeoFus}, a task-adaptive geometric fusion module that uses text-and-scene-conditioned gating to select among shallow, intermediate, and deep geometric features, and position-specific projection and residual injection to integrate them across LLM decoder layers for reasoning.

\subsubsection{Multi-Level Geometric Projection.}
Let $s\in\{0,1,2\}$ denote a geometric feature injection position in the LLM, and let $\mathbf{V}_s\in\mathbb{R}^{(KN')\times D_{\mathrm{lang}}}$ represent the hidden states of the visual tokens at that position. For each candidate geometric layer $\ell\in\mathcal{K}$, we introduce a position- and layer-specific projector $P_{s,\ell}$ to adapt the geometric features to the LLM hidden representation at injection position $s$:
$\mathbf{Z}_{s,\ell}=P_{s,\ell}(\mathbf{G}_{\ell})$.
Each projector consists of LayerNorm followed by a two-layer MLP:
\begin{equation}
P_{s,\ell}(\mathbf{G}_{\ell})
=
\mathbf{W}^{(2)}_{s,\ell}
\sigma\left(
\mathbf{W}^{(1)}_{s,\ell}
\operatorname{LN}(\mathbf{G}_{\ell})
+
\mathbf{b}^{(1)}_{s,\ell}
\right)
+
\mathbf{b}^{(2)}_{s,\ell}.
\end{equation}
Here, $\mathbf{W}^{(1)}_{s,\ell}$ and $\mathbf{W}^{(2)}_{s,\ell}$ are learnable weight matrices, $\mathbf{b}^{(1)}_{s,\ell}$ and $\mathbf{b}^{(2)}_{s,\ell}$ are learnable bias terms, and $\sigma(\cdot)$ denotes the activation function. By decoupling the projection parameters across injection positions and geometric layers, the model can learn position-specific transformations for the same geometric representation at different levels of the LLM.

\begin{table*}[t]
    
    \caption{\small{
    \textbf{Evaluation across spatial intelligence benchmarks.}
    \bestscore{Bold} and \secondscore{underlined} values indicate the best
    and second-best results within each open-source spatial intelligence
    model scale group, respectively.
    \textsuperscript{\(\dagger\)} denotes models trained using the same
    training data.
    }}
    \label{tab:spatial_benchmark_results}

    \begingroup
    \renewcommand{\arraystretch}{0.95}
    \setlength{\tabcolsep}{2.8pt}
    \setlength{\aboverulesep}{1.4pt}
    \setlength{\belowrulesep}{1.4pt}
    \fontsize{5.3}{6.5}\selectfont
    \setlength{\abovecaptionskip}{1pt}

    \resizebox{\textwidth}{!}{%
    \begin{tabular}{
        >{\raggedright\arraybackslash}p{4.8cm}
        c
        c
        *{6}{c}
    }

        \toprule

        \textbf{Methods}
        & \textbf{Rank}
        & \textbf{Avg.}
        & \textbf{SparBench}
        & \textbf{VSI-Bench}
        & \textbf{SPBench}
        & \textbf{BLINK}
        & \textbf{MindCube}
        & \shortstack{\textbf{SQA3D}}
        \\

        \midrule

        \rowcolor{sectionblue}
        \multicolumn{9}{l}{
            \textbf{\textit{Proprietary Models (API)}}
        }
        \\

        GPT-5.6-Luna
        & 3 & 52.4 & 47.7 & 52.2 & 54.2 & 65.6 & 56.4 & 38.1 \\

        Doubao-Seed-2.0-Pro
        & 2 & 54.1 & 55.5 & 49.3 & 50.3 & 50.6 & 61.4 & 57.7 \\

        Gemini-3.5-Flash
        & 1 & 54.3 & 57.2 & 56.5 & 61.8 & 33.9 & 71.6 & 45.0 \\

        \midrule

        \rowcolor{sectionblue}
        \multicolumn{9}{l}{
            \textbf{\textit{Open-Source Models}}
        }
        \\

        Qwen2.5-VL-3B
        &  & 39.1 & 29.5 & 27.4 & 40.5 & 48.9 & 41.8 & 46.6 \\

        Qwen2.5-VL-7B
        &  & 40.3 & 33.0 & 31.5 & 43.3 & 53.7 & 34.7 & 45.7 \\

        Qwen3.5-4B
        &  & 44.9 & 39.8 & 50.1 & 53.2 & 29.5 & 48.7 & 48.1 \\

        Qwen3-VL-4B
        & 3 & 45.5 & 36.3 & 54.6 & 57.6 & 47.4 & 25.4 & 51.6 \\

        Qwen3.5-9B
        & 2 & 49.0 & 46.7 & 55.7 & 53.2 & 42.5 & 43.3 & 52.6 \\

        Qwen3.5-4B\textsuperscript{\(\dagger\)}
        & 1 & 67.0 & 68.5 & 66.5 & 67.5 & 57.9 & 78.9 & 62.9 \\
        
        \midrule

        \rowcolor{sectionblue}
        \multicolumn{9}{l}{
            \textbf{\textit{
                Specialized Spatial Intelligence Models (3--5B)
            }}
        }
        \\

        VG-LLM-4B
        &  & 30.8
        & 20.7
        & 47.3
        & 32.2
        & 38.2
        & 33.7
        & 12.6 \\

        Spatial-MLLM-4B
        &  & 45.5
        & 32.5
        & 50.6
        & 54.7
        & 41.5
        & 38.9
        & \secondscore{54.7} \\

        SpatialStack-5B
        & 3 & 57.5
        & 68.3
        & 67.1
        & 68.7
        & \bestscore{59.1}
        & 41.5
        & 40.1 \\

        SpatialStack-5B\textsuperscript{\(\dagger\)}
        & 2 & \secondscore{68.6}
        & \secondscore{69.7}
        & \secondscore{69.6}
        & \secondscore{69.9}
        & 54.2
        & \secondscore{94.5}
        & 53.8 \\

        \textbf{SpatialMind-5B (Ours)}
        & 1 & \textbf{70.7}
        & \bestscore{70.0}
        & \bestscore{71.5}
        & \bestscore{71.6}
        & \secondscore{54.3}
        & \bestscore{94.6}
        & \bestscore{62.2} \\

        \midrule

        \rowcolor{sectionblue}
        \multicolumn{9}{l}{
            \textbf{\textit{
                Specialized Spatial Intelligence Models (8--10B)
            }}
        }
        \\

        SpatialStack-10B
        &  & 58.3
        & \secondscore{69.7}
        & 68.8
        & 68.2
        & \bestscore{64.3}
        & 33.8
        & 44.8 \\

        GeoThinker-8B (Qwen2.5-VL-7B)
        &  & 60.9
        & 68.4
        & 67.9
        & 56.6
        & 48.9
        & 91.3
        & 32.1 \\

        GeoThinker-9B (Qwen3-VL-8B)
        & 3 & 63.2
        & 69.1
        & \secondscore{71.9}
        & 61.2
        & 53.9
        & 90.4
        & 32.9 \\

        SpatialStack-10B\textsuperscript{\(\dagger\)}
        & 2 & \secondscore{70.1}
        & 68.8
        & 71.3
        & \bestscore{69.0}
        & 56.1
        & \secondscore{91.7}
        & \secondscore{63.5} \\

        \textbf{SpatialMind-10B (Ours)}
        & 1 & \textbf{71.6}
        & \bestscore{70.3}
        & \bestscore{72.3}
        & \secondscore{68.4}
        & \secondscore{60.0}
        & \bestscore{93.9}
        & \bestscore{64.5} \\

        \bottomrule
        
    \end{tabular}%
    }

    \endgroup
\end{table*}

\subsubsection{Text-and-Scene-Conditioned Gating.}
To make multi-level geometric fusion adaptive to both query semantics and scene content, we construct a unified gating condition from textual, visual, and geometric representations. At the $s$-th injection position, the textual and visual conditions are defined as
$\mathbf{c}_{t}=\operatorname{MaskedAvgPool}(\mathbf{T})$
and
$\mathbf{c}_{v,s}=\operatorname{AvgPool}(\mathbf{V}_{s})$,
respectively. Here, $\operatorname{MaskedAvgPool}$ aggregates only the tokens corresponding to the input query while excluding padding and target-answer tokens.

To incorporate information from all candidate geometric layers, we first compute a global descriptor for each projected geometric representation:
$\mathbf{g}_{s,\ell}
=
\operatorname{AvgPool}(\mathbf{Z}_{s,\ell})
\in
\mathbb{R}^{D_{\mathrm{lang}}}$,
where $\ell\in\mathcal{K}$. These descriptors are then aggregated into a unified geometric condition:
\begin{equation}
\mathbf{c}_{g,s}
=
\frac{1}{|\mathcal{K}|}
\sum_{\ell\in\mathcal{K}}
\operatorname{LN}_{g,\ell}(\mathbf{g}_{s,\ell}).
\end{equation}
The three conditioning representations are combined to form the gating condition:
\begin{equation}
\mathbf{c}_s
=
\operatorname{LN}_{t}(\mathbf{c}_t)
+
\alpha_{v,s}\operatorname{LN}_{v}(\mathbf{c}_{v,s})
+
\alpha_{g,s}\operatorname{LN}_{g}(\mathbf{c}_{g,s}),
\end{equation}
where $\alpha_{v,s}$ and $\alpha_{g,s}$ are learnable scaling factors. Based on $\mathbf{c}_s$, the gating network predicts the selection logits
$\mathbf{a}_s=\mathbf{W}_s\mathbf{c}_s+\mathbf{b}_s\in\mathbb{R}^{|\mathcal{K}|}$.
The logits are converted into normalized fusion weights using a temperature-controlled softmax:
\begin{equation}
w_{s,\ell}
=
\frac{\exp(a_{s,\ell}/\tau)}
{\sum_{j\in\mathcal{K}}\exp(a_{s,j}/\tau)},
\qquad \ell\in\mathcal{K},
\end{equation}
where $\tau>0$ is the temperature parameter. The resulting weights satisfy
$w_{s,\ell}\geq0$ and
$\sum_{\ell\in\mathcal{K}}w_{s,\ell}=1$.
They assess the relevance of each geometric level to the query and scene at position $s$, enabling adaptive multi-level feature fusion.

\subsubsection{Feature Fusion and Residual Injection.}
Given the predicted gating weights, we construct the mixed geometric representation as a convex combination of the projected geometric features:
\begin{equation}
\widetilde{\mathbf{G}}_s
=
\sum_{\ell\in\mathcal{K}}
w_{s,\ell}\mathbf{Z}_{s,\ell}.
\end{equation}
We then inject the mixed geometric representation into the visual-token hidden states through a residual connection:
\begin{equation}
\widetilde{\mathbf{V}}_s
=
\mathbf{V}_s+\widetilde{\mathbf{G}}_s.
\end{equation}
This residual formulation preserves a direct pathway for the original visual information while incorporating the adaptively fused geometric representation.

\subsection{Negative-Entropy Regularization}
During training, the gating network may concentrate most of its weight on a single geometric layer, preventing the model from fully exploiting the complementary information encoded at different geometric levels. To mitigate this gating-collapse problem, we introduce a negative-entropy regularization term:
\begin{equation}
\mathcal{L}_{\mathrm{gate}}
=
-\lambda \frac{1}{S}
\sum_{s=0}^{S-1} H(\mathbf{w}_s),
\end{equation}
where $S=3$ denotes the number of geometric injection positions, $\lambda=0.005$ controls the regularization strength, and
$\mathbf{w}_s=\{w_{s,\ell}\}_{\ell\in\mathcal{K}}$
is the gating distribution at the $s$-th injection position. Its Shannon entropy is defined as
$H(\mathbf{w}_s)
=
-\sum_{\ell\in\mathcal{K}}
w_{s,\ell}\log w_{s,\ell}$.
When the gating distribution collapses onto a single geometric layer,
$H(\mathbf{w}_s)$ approaches zero. In contrast, the entropy increases as the distribution becomes more balanced and reaches its maximum value
$\log|\mathcal{K}|$ under a uniform distribution. Therefore, minimizing
$\mathcal{L}_{\mathrm{gate}}$ discourages gating collapse without preventing subsequent task-dependent geometric specialization. In the second training stage, the overall objective is defined as
$\mathcal{L}
=
\mathcal{L}_{\mathrm{LM}}
+
\mathcal{L}_{\mathrm{gate}}$,
where $\mathcal{L}_{\mathrm{LM}}$ denotes the standard autoregressive language modeling loss.

\section{Experiments}
Sec.~5.1 details the training setup, Sec.~5.2 reports the results on spatial reasoning benchmarks, Sec.~5.3 presents component ablation studies, and Sec.~5.4 provides visualization of the adaptive gating weights.
\subsection{Training}
\subsubsection{Training Data Construction.}
To cover diverse spatial reasoning tasks, we construct the training set
through balanced sampling across multiple spatial reasoning datasets.
Specifically, the training corpus combines VG-LLM \cite{zheng2026learning}, including the SPAR and LLaVA-Hound subsets, the ScanNet subset from VLM-3R \cite{fan2026vlm}, a curated subset of VSI-590K \cite{yang2025cambrian}, ViCA-322K \cite{feng2025visuospatial}, MindCube \cite{yin2025spatial}, and Spatial-MLLM-203K \cite{wu2026spatial}. By combining these complementary data sources, we enhance the model's spatial perception and
reasoning capabilities in complex 3D environments.
 
\subsubsection{Two-Stage Training Strategy.}
To improve optimization stability, we adopt a two-stage training strategy. We use Qwen3.5-4B \cite{qwen35blog} as the multimodal backbone and a frozen VGGT-1B \cite{wang2025vggt} to provide geometric priors. The native visual encoder of Qwen3.5-4B and VGGT remain frozen throughout both training stages.
\textbf{In Stage 1,} we adopt the fixed layer-wise mapping of SpatialStack \cite{zhang2026spatialstack} and jointly optimize the LLM and geometric projectors for one epoch using the standard autoregressive language modeling objective. We use AdamW with a learning rate of $1\times10^{-5}$, a warmup ratio of $0.03$, a cosine learning-rate schedule, and a global batch size of $64$.
\textbf{In Stage 2,} GeoFus is activated for adaptive multi-level geometric fusion. The LLM backbone and language modeling head are frozen, while only the geometric projectors $P_{s,\ell}$, gating parameters $(\mathbf{W}_s,\mathbf{b}_s)$, and scaling factors $\alpha_{v,s}$ and $\alpha_{g,s}$ are optimized. In Stage 2, we optimize $\mathcal{L}_{\mathrm{LM}} + \mathcal{L}_{\mathrm{gate}}$
using AdamW with a learning rate of $5 \times 10^{-5}$, a global batch size of
64, and a cosine learning-rate schedule for 1 epoch. All experiments are conducted on 8 NVIDIA A800 GPUs. Training SpatialMind-5B takes approximately 30 hours in Stage 1 and 15 hours in Stage 2.

\subsection{Main Results}
\subsubsection{Evaluation on VSI-Bench. }VSI-Bench \cite{yang2025thinking} comprises over 5,000 carefully curated questions derived from the ARKitScenes \cite{baruch2021arkitscenes}, ScanNet \cite{dai2017scannet}, and ScanNet++ \cite{yeshwanth2023scannet++}. It includes two answer formats, Numerical Answers (NA) and Multiple-Choice Answers (MCA). Following the official evaluation protocol, we use relative accuracy for NA questions and mean accuracy for MCA questions. We compare SpatialMind with proprietary models, open-source MLLMs, and specialized spatial reasoning models in the 3–5B and 8–10B parameter groups. As shown in Table~\ref{tab:vsibench_main_results}, SpatialMind-5B achieves a state-of-the-art overall score of 71.5 among models in the 3--5B parameter range, while SpatialMind-10B reaches 72.3 and establishes a new state of the art among 8--10B models. The consistent gains across both parameter ranges indicate that the improvements are not limited to a specific backbone scale, but arise from the effective adaptive fusion of multi-level geometric representations.

\subsubsection{Evaluation on Other Spatial Understanding Benchmarks.}
To assess the overall spatial understanding capability of SpatialMind, we further evaluate it on five additional spatial understanding benchmarks: SPAR-Bench \cite{zhang2026flatland}, SPBench \cite{stogiannidis2025mind}, BLINK \cite{fu2024blink}, MindCube \cite{yin2025spatial}, and SQA3D \cite{ma2022sqa3d}. These benchmarks cover diverse task formats and spatial capabilities, ranging from multi-view and single-image spatial reasoning to 3D question answering. As shown in Table~\ref{tab:spatial_benchmark_results}, SpatialMind-5B achieves the highest average score of 70.7 in the 3--5B group, outperforming the same-data SpatialStack-5B by 2.1 points and ranking first on five of the six benchmarks. SpatialMind-10B achieves the highest average score of 71.6 in the 8–10B group and ranks first on four of the six benchmarks. These results demonstrate that SpatialMind delivers strong performance across diverse spatial intelligence benchmarks.

\begin{table}[t]
    
    \caption{\small{\textbf{Ablation Study of Components on VSI-Bench.} Checkmarks indicate enabled components. The complete GeoFus configuration achieves the best average score of $71.5\%$.}}
    \label{tab:ablation}

    \begingroup
    \renewcommand{\arraystretch}{1.05}
    \setlength{\tabcolsep}{2.2pt}
    \scriptsize

    \resizebox{\columnwidth}{!}{%
    \begin{tabular}{
        c c c c
        c
        *{4}{c}
        *{4}{c}
    }
        \toprule

        \multicolumn{2}{c}{\textbf{Gating Conditions}}
        & \multirow{2}{*}{\makecell{\textbf{Cross-Layer}\\\textbf{Selection}}}
        & \multirow{2}{*}{\makecell{\textbf{Neg.-Entropy}\\\textbf{Reg.}}}
        & \multirow{2}{*}{Avg.}
        & \multicolumn{4}{c}{Numerical Answer}
        & \multicolumn{4}{c}{Multiple-Choice Answer}   
        \\

        \cmidrule(lr){1-2}
        \cmidrule(lr){6-9}
        \cmidrule(lr){10-13}

        \textbf{Text}
        & \makecell{\textbf{Semantic}\\\textbf{+ Spatial}}
        & &
        &
        &
        \makecell{Obj.\\Count}
        & \makecell{Abs.\\Dist.}
        & \makecell{Obj.\\Size}
        & \makecell{Room\\Size}
        & \makecell{Rel.\\Dist.}
        & \makecell{Rel.\\Dir.}
        & \makecell{Route\\Plan}
        & \makecell{Appr.\\Order}
        \\

        \midrule

        \no
        & \no
        & \no
        & \no
        & 69.6
        & 70.7
        & 60.6
        & 76.7
        & 73.1
        & 66.3
        & 88.9
        & 46.9
        & 73.3
        \\

        \yes
        & \no
        & \no
        & \no
        & 70.1
        & 71.5
        & 60.2
        & 76.8
        & 72.5
        & 68.9
        & 86.2
        & 43.8
        & 81.4
        \\

        \yes
        & \yes
        & \no
        & \no
        & 70.2
        & 71.0
        & 61.2
        & 77.1
        & 73.4
        & 69.0
        & 84.9
        & 46.4
        & 78.8
        \\

        \yes
        & \yes
        & \yes
        & \no
        & 70.9
        & 69.8
        & 65.2
        & 76.5
        & 76.9
        & 70.3
        & 91.5
        & 44.8
        & 72.3
        \\

        \yes
        & \yes
        & \yes
        & \yes
        & \textbf{71.5}
        & 71.8
        & 65.0
        & 77.1
        & 77.7
        & 70.7
        & 91.8
        & 45.9
        & 72.0
        \\

        \bottomrule
    \end{tabular}%
    }

    \endgroup
\end{table}

\subsection{Ablation Studies}
To assess the contribution of each GeoFus component, we perform cumulative ablations using SpatialMind-5B, with results averaged over three independent runs on VSI-Bench. As shown in Table~\ref{tab:ablation}, the SpatialStack-style baseline achieves $69.6\%$, and introducing a text-conditioned scalar gate while retaining the fixed layer mappings improves the score to $70.1\%$. Incorporating joint textual, visual-semantic, and spatial conditions further increases the score to $70.2\%$.  Enabling each injection position to dynamically select and combine multiple geometric layers improves the score to $70.9\%$, representing the largest incremental gain and demonstrating the importance of cross-layer selection. Negative-entropy regularization further raises the score to $71.5\%$, yielding an overall improvement of $1.9$ percentage points over the fixed baseline. Overall, cross-layer dynamic selection provides the primary performance gain, while conditioning and negative-entropy regularization further enhance adaptive geometric fusion.

\begin{figure}[t]
    
    \captionsetup{skip=0pt} 
    \includegraphics[width=\columnwidth]{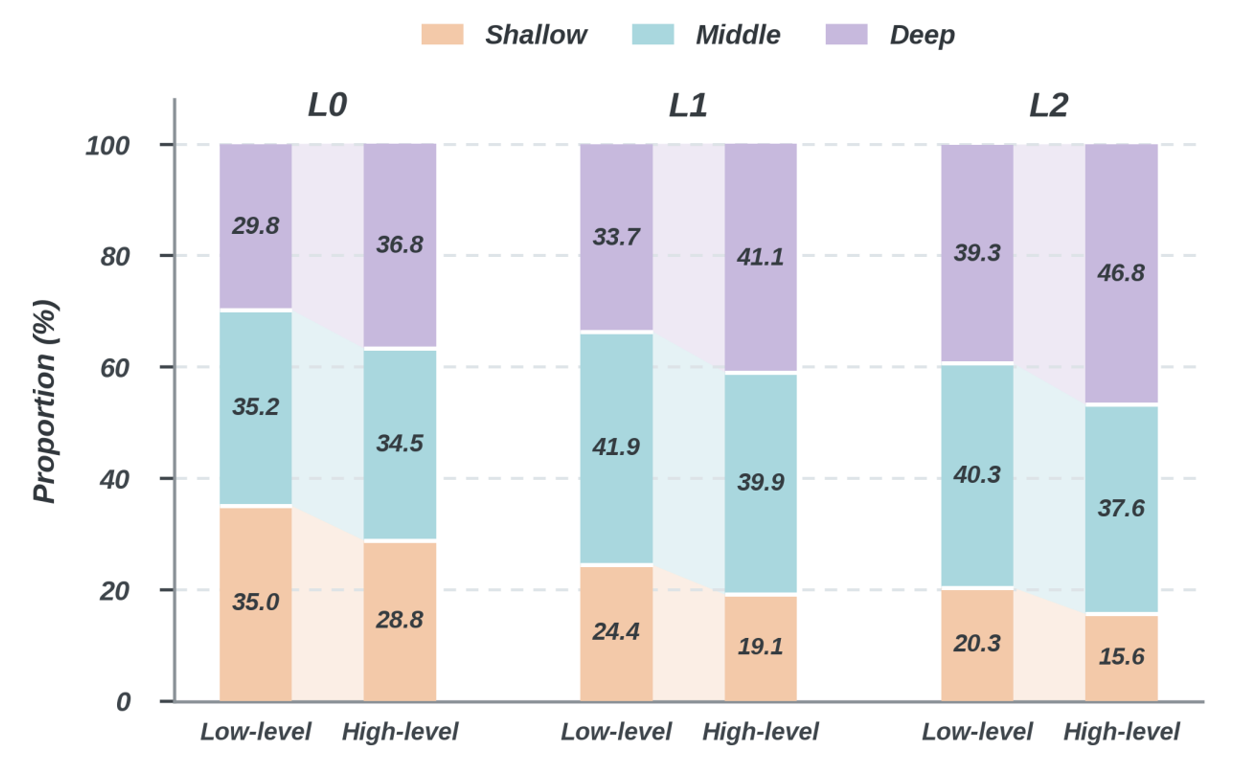} 
    \caption{{\small{\textbf{Visualization of Adaptive Gating Weights.} Across the L0--L2 injection layers, low-level tasks favor shallow geometric features, whereas high-level tasks emphasize deep features, demonstrating task-adaptive selection of geometric representations.}}}
    \label{fig:Visualization of Adaptive Gating Weights}
\end{figure}

\subsection{Visualization of the Adaptive Gating Weights}
To examine how SpatialMind adaptively selects multi-level geometric features, we randomly sample 50 examples from each of the eight VSI-Bench subtasks. We first average the learned gating weights within each subtask and then macro-average them over the low-level perception and high-level reasoning groups. As shown in Fig.~\ref{fig:Visualization of Adaptive Gating Weights}, low-level perception tasks assign larger weights to shallow features than high-level reasoning tasks at all three injection positions (L0: $35.0\%$ vs.\ $28.8\%$; L1: $24.4\%$ vs.\ $19.1\%$; L2: $20.3\%$ vs.\ $15.6\%$). Conversely, high-level reasoning tasks place greater emphasis on deep features than low-level perception tasks (L0: $36.8\%$ vs.\ $29.8\%$; L1: $41.1\%$ vs.\ $33.7\%$; L2: $46.8\%$ vs.\ $39.3\%$). Overall, as task complexity increases, SpatialMind gradually reduces its reliance on shallow geometric cues while strengthening the contribution of deep representations, with intermediate features maintaining a relatively stable role. These results demonstrate that GeoFus can dynamically adjust geometric granularity according to task requirements.

\section{Conclusion}
In this work, we introduced \textit{SpatialMind}, a task-adaptive framework for multi-level geometric fusion in MLLMs. Our analysis reveals that geometric features exhibit hierarchical specialization and cross-level complementarity: shallow features preserve fine-grained spatial details, while deeper features capture global scene structures and high-level relations. This finding also explains why fixed fusion strategies are inadequate, as diverse spatial tasks require geometric cues at different levels of granularity. To address this limitation, we developed GeoFus, which jointly leverages textual, visual, and geometric cues to adaptively fuse multi-level geometric features across LLM layers. Negative-entropy regularization is further introduced to mitigate gate collapse. Visualizations of the learned gating weights show that GeoFus selects geometric representations at task-appropriate granularities. Our experiments demonstrate the strong performance of SpatialMind across diverse spatial reasoning benchmarks. On VSI-Bench, SpatialMind-5B and SpatialMind-10B achieve state-of-the-art scores of 71.5 and 72.3 within the 3--5B and 8--10B parameter ranges, respectively, validating task-adaptive geometric fusion across model scales.

\bibliography{aaai2027}

\clearpage
\appendix
\section*{Supplementary Material Overview}

\noindent
This supplementary material is organized into four sections.

\noindent\textbf{Section 1} provides the model configuration, two-stage training strategy, optimization settings, and gate initialization.

\noindent\textbf{Section 2} describes the training-data construction and the evaluation protocols and results on additional spatial reasoning benchmarks.

\noindent\textbf{Section 3} reports the complete multi-level geometric ablation results and analyzes the specialization and complementarity of shallow, intermediate, and deep features.

\noindent\textbf{Section 4} presents the VSI-Bench task taxonomy, per-task gating analysis, task-dependent response visualization, and representative qualitative results.

\section{Implementation Details}

\subsection{Model Configuration}

We instantiate SpatialMind at two model scales. SpatialMind-5B and
SpatialMind-10B employ Qwen3.5-4B and Qwen3.5-9B as their multimodal
backbones, respectively, together with a frozen VGGT-1B geometry encoder.
The native vision encoder and visual MLP projector are kept frozen throughout
training. We extract multi-level geometric representations from VGGT layers
11, 17, and 23, which respectively encode shallow, intermediate, and deep
geometric cues. These representations are integrated at LLM layers 0, 1, and
2.

\subsection{Two-Stage Training Strategy}

We adopt a two-stage training strategy to stabilize the optimization of
adaptive geometric fusion.
\textbf{In Stage~1}, we follow a fixed layer-wise mapping:
VGGT layer 11 is injected into LLM layer 0, VGGT layer 17 into LLM layer 1,
and VGGT layer 23 into LLM layer 2. The LLM and geometric projectors are
jointly optimized using the standard autoregressive language-modeling
objective, while VGGT, the native vision encoder, and the visual MLP projector
remain frozen. This stage establishes a stable correspondence between
multi-level geometric representations and the early LLM layers.
\textbf{In Stage~2}, training is initialized from the Stage-1 checkpoint and GeoFus is
activated. At each injection position, GeoFus adaptively combines geometric
representations from VGGT layers 11, 17, and 23. The LLM backbone, language
modeling head, VGGT, native vision encoder, and visual MLP projector are
frozen. Only the position- and layer-specific geometric projectors
$P_{s,k}$, gating parameters $(W_s,b_s)$, and scaling factors
$\alpha_{v,s}$ and $\alpha_{g,s}$ are optimized. The Stage-2 objective is
\begin{equation}
    \mathcal{L}
    =
    \mathcal{L}_{\mathrm{LM}}
    +
    \mathcal{L}_{\mathrm{gate}},
\end{equation}
where $\mathcal{L}_{\mathrm{gate}}$ is the negative-entropy regularization
term used to mitigate gate collapse.

\subsection{Optimization Settings}

Both stages are trained for one epoch using AdamW with BF16 precision, a
global batch size of 64, a cosine learning-rate schedule, a warmup ratio of
0.03, and a weight decay of 0.01. The learning rates are
$1\times10^{-5}$ for Stage~1 and $5\times10^{-5}$ for Stage~2. We employ
DeepSpeed ZeRO-2 and gradient checkpointing for distributed training. All
experiments are conducted on eight NVIDIA A800 80GB GPUs. For
SpatialMind-5B, Stage~1 and Stage~2 take approximately 30 and 15 hours,
respectively. The detailed configurations are summarized in
Table~\ref{tab:training_details}.
\begin{table*}[t]
\centering
\caption{\textbf{Training configurations for the two-stage optimization strategy.}}
\label{tab:training_details}
\small
\setlength{\tabcolsep}{5pt}
\renewcommand{\arraystretch}{1.08}

\begin{tabularx}{\textwidth}{@{}
    >{\raggedright\arraybackslash}p{0.18\textwidth}
    >{\raggedright\arraybackslash}p{0.38\textwidth}
    >{\raggedright\arraybackslash}X
@{}}
\toprule
\textbf{Parameter}
& \textbf{Stage 1: Fixed-Fusion Training}
& \textbf{Stage 2: Adaptive GeoFus Training} \\
\midrule

Base checkpoint
& Qwen3.5-4B / Qwen3.5-9B
& Stage-1 fixed-fusion checkpoint \\

Geometry encoder
& VGGT-1B
& VGGT-1B \\

Trainable modules
& LLM and geometry-fusion modules
& GeoFus modules only \\

\midrule

Optimizer and precision
& AdamW, BF16
& AdamW, BF16 \\

Learning rate
& $1\times10^{-5}$
& $5\times10^{-5}$ \\

Epochs
& 1
& 1 \\

Global batch size
& 64
& 64 \\

\midrule

Fusion method
& Fixed layer-wise geometric injection
& Text- and scene-conditioned adaptive GeoFus \\

VGGT-to-LLM mapping
& \makecell[tl]{
    VGGT-11 to LLM-0 \\
    VGGT-17 to LLM-1 \\
    VGGT-23 to LLM-2
}
& \makecell[tl]{
    Adaptive selection among \\
    VGGT-11, VGGT-17, and VGGT-23 \\
    at each LLM fusion layer
} \\

Gate type
& Disabled
& Three-way softmax mixture \\

Temperature
& N/A
& $\tau=1.0$ \\

Gate regularization
& N/A
& Negative entropy, $\lambda=0.005$ \\

Gate conditioning
& N/A
& Textual, visual-scene, and geometry-scene signals \\

\bottomrule
\end{tabularx}
\end{table*}
\subsection{Gate Initialization}

To preserve the inductive bias learned through the fixed mapping in Stage~1,
we initialize each gate to favor its originally assigned geometric layer. Let
$k_s$ denote the original VGGT layer associated with injection position $s$,
where $k_0=11$, $k_1=17$, and $k_2=23$. The gating weight matrix is
initialized as $W_s=0$, while the bias is initialized as
\begin{equation}
    b_{s,k}
    =
    \begin{cases}
        1,  & k=k_s,\\
        -1, & k\neq k_s.
    \end{cases}
\end{equation}
With a softmax temperature of $\tau=1$, the initial weight of the original
geometric pathway is
\begin{equation}
    w_{s,k_s}
    =
    \frac{\exp(1)}
    {\exp(1)+2\exp(-1)}
    \approx 0.787,
\end{equation}
whereas the weight of each alternative pathway is
\begin{equation}
    w_{s,k}
    =
    \frac{\exp(-1)}
    {\exp(1)+2\exp(-1)}
    \approx 0.106,
    \qquad k\neq k_s.
\end{equation}
Thus, the initial gating distribution is approximately
$(0.8,0.1,0.1)$ after arranging the original pathway first. This
initialization preserves the Stage-1 fusion pattern at the beginning of
Stage~2 while allowing the gate to subsequently learn task-dependent
geometric specialization. The negative-entropy regularization term is
defined as
\begin{equation}
    \mathcal{L}_{\mathrm{gate}}
    =
    -\lambda\frac{1}{S}
    \sum_{s=1}^{S}H(\mathbf{w}_s),
\end{equation}
where $S=3$ and $\lambda=0.005$.

\section{Dataset Details}

\subsection{Training Data Construction}

To cover diverse spatial perception and reasoning tasks, we construct the
training corpus through balanced sampling from multiple spatial
instruction-tuning datasets. Specifically, SPAR provides broad hierarchical
supervision spanning fine-grained geometric perception, viewpoint
transformation, and high-level scene reasoning, while LLaVA-Hound preserves
general video understanding and instruction-following capabilities. The
ScanNet subset of VLM-3R provides reconstruction-aligned supervision for
recovering 3D scene structure, cross-view correspondences, and object-level
spatial relations from monocular observations. A curated subset of VSI-590K
emphasizes video-centric indoor spatial skills, including distance and
direction estimation, object- and room-scale perception, route reasoning,
counting, and appearance-order understanding. ViCA-322K complements these
sources with metadata-grounded question--answer pairs from real-world indoor
videos, strengthening multi-view scene integration and complex video-based
spatial reasoning. MindCube focuses on spatial mental modeling from limited
views, providing supervision for cognitive mapping, perspective taking, and
mental simulation. Finally, Spatial-MLLM-203K contributes scene-level and
situated spatial question answering, improving reasoning over implicit 3D
scene structures from visual observations. For each source, we use its designated training split
and evaluate SpatialMind on the official benchmark split.

\subsection{Evaluation Benchmarks and Metrics}
To assess the overall spatial understanding capability of SpatialMind, we further evaluate it on five additional spatial understanding benchmarks: SPAR-Bench \cite{zhang2026flatland}, SPBench \cite{stogiannidis2025mind}, BLINK \cite{fu2024blink}, MindCube \cite{yin2025spatial}, and SQA3D \cite{ma2022sqa3d}. These benchmarks cover diverse task formats and spatial capabilities, ranging from multi-view and single-image spatial reasoning to 3D question answering.

\noindent\textbf{MindCube.}
MindCube contains 21,154 questions over 3,268 images
and evaluates whether a model can construct a spatial mental model from
limited visual observations. It covers three core capabilities: cognitive
mapping, perspective taking, and mental simulation under hypothetical
movements. We follow the official multiple-choice evaluation protocol and
report overall accuracy. Detailed results for the Among, Around, and Rotation
tasks are reported in Table~\ref{tab:mindcubebench}.

\begin{table}[!t]
    \centering
    \caption{\small{
        \textbf{Evaluation on MindCubeBench.}
        \bestscore{Bold} and \secondscore{underlined} values indicate the
        best and second-best results within each specialized-model size group,
        respectively. \textsuperscript{\(\dagger\)} denotes models trained
        using the same training data.
    }}
    \label{tab:mindcubebench}

    \begingroup
    \renewcommand{\arraystretch}{0.95}
    \setlength{\tabcolsep}{1.0pt}
    \setlength{\aboverulesep}{1.0pt}
    \setlength{\belowrulesep}{1.0pt}
    \fontsize{6.2}{6.9}\selectfont
    
    \begin{tabularx}{\columnwidth}{@{}
    >{\raggedright\arraybackslash}X
    >{\centering\arraybackslash}p{0.065\columnwidth}
    >{\centering\arraybackslash}p{0.095\columnwidth}
    >{\centering\arraybackslash}p{0.090\columnwidth}
    >{\centering\arraybackslash}p{0.090\columnwidth}
    >{\centering\arraybackslash}p{0.090\columnwidth}
@{}}
        \toprule
        \textbf{Methods}
        & \textbf{Rank}
        & \textbf{Overall}
        & \textbf{Among}
        & \textbf{Around}
        & \textbf{Rotation} \\
        \midrule

        \rowcolor{sectionblue}
        \multicolumn{6}{l}{\textbf{\textit{Proprietary Models (API)}}} \\
        GPT-5.6-Luna & 3 & 56.44 & 41.69 & 72.40 & 80.00 \\
        Doubao-Seed-2.0-Pro & 2 & 61.35 & 47.29 & 67.20 & 95.50 \\
        Gemini-3.5-Flash & 1 & 71.63 & 61.53 & 79.20 & 92.00 \\

        \midrule
        \rowcolor{sectionblue}
        \multicolumn{6}{l}{\textbf{\textit{Open-Source Models}}} \\
        Qwen2.5-VL-3B-Instruct &  & 41.83 & 42.20 & 48.40 & 32.50 \\
        Qwen2.5-VL-7B-Instruct &  & 34.71 & 32.54 & 40.00 & 34.50 \\
        Qwen3.5-4B & 2 & 48.65 & 43.56 & 60.80 & 48.50 \\
        Qwen3-VL-4B-Instruct &  & 25.38 & 18.64 & 32.80 & 36.00 \\
        Qwen3.5-9B & 3 & 43.27 & 41.53 & 58.40 & 29.50 \\
        Qwen3.5-4B\textsuperscript{\(\dagger\)}
        & 1 & 78.85 & 80.00 & 73.60 & 82.00 \\

        \midrule
        \rowcolor{sectionblue}
        \multicolumn{6}{l}{
            \textbf{\textit{Open-Source Spatial Intelligence Models (3--5B)}}
        } \\
        VG-LLM-4B &  & 33.65 & 33.22 & 38.40 & 29.00 \\
        Spatial-MLLM-4B &  & 38.85 & 35.76 & 50.40 & 33.50 \\
        SpatialStack-5B & 3 & 41.54 & 41.02 & 45.20 & 38.50 \\
        SpatialStack-5B\textsuperscript{\(\dagger\)}
        & 2
        & \secondscore{94.52}
        & \bestscore{96.61}
        & \bestscore{94.40}
        & \secondscore{88.50} \\
        \textbf{SpatialMind-5B (ours)}
        & 1
        & \bestscore{94.62}
        & \secondscore{96.44}
        & \secondscore{94.00}
        & \bestscore{90.00} \\

        \midrule
        \rowcolor{sectionblue}
        \multicolumn{6}{l}{
            \textbf{\textit{Open-Source Spatial Intelligence Models (7--10B)}}
        } \\
        SpatialStack-10B &  & 33.75 & 30.68 & 35.60 & 40.50 \\
        GeoThinker (Qwen2.5-VL-7B)
        & 3 & 91.25 & 93.90 & 89.60 & \bestscore{85.50} \\
        GeoThinker (Qwen3-VL-8B)
        &  & 90.38 & 95.59 & 86.00 & 80.50 \\
        SpatialStack-10B\textsuperscript{\(\dagger\)}
        & 2
        & \secondscore{91.73}
        & \secondscore{96.78}
        & \bestscore{94.40}
        & 73.50 \\
        \textbf{SpatialMind-10B (ours)}
        & 1
        & \bestscore{93.85}
        & \bestscore{97.29}
        & \secondscore{92.80}
        & \secondscore{85.00} \\

        \bottomrule
    \end{tabularx}
    \endgroup
\end{table}

\begin{table*}[!t]
    \centering
    \caption{\small{
        \textbf{Evaluation on SQA3D.}
        \bestscore{Bold} and \secondscore{underlined} values indicate the
        best and second-best results within each specialized-model size group,
        respectively. \textsuperscript{\(\dagger\)} denotes models trained
        using the same training data.
    }}

    \begingroup
    \renewcommand{\arraystretch}{0.92}
    \setlength{\tabcolsep}{2.6pt}
    \setlength{\aboverulesep}{0.7pt}
    \setlength{\belowrulesep}{0.7pt}
    \fontsize{4.4}{5.0}\selectfont

    \resizebox{0.97\textwidth}{!}{%
    \begin{tabular}{
        >{\raggedright\arraybackslash}p{4.8cm}
        cccccccc
    }
        \toprule
        \multirow{2}{*}{\textbf{Methods}}
        & \multirow{2}{*}{\textbf{Rank}}
        & \multirow{2}{*}{\textbf{Avg. EM}}
        & \multicolumn{6}{c}{
            \cellcolor{headercream}\textbf{EM by Question Type}
        } \\
        \cmidrule(lr){4-9}
        & & & What & Is & How & Can & Which & Others \\
        \midrule

        \rowcolor{sectionblue}
        \multicolumn{9}{l}{\textbf{\textit{Open-Source Models}}} \\
        Qwen2.5-VL-3B-Instruct &  & 46.43 & 40.28 & 53.53 & 43.44 & 53.25 & 47.29 & 48.59 \\
        Qwen2.5-VL-7B-Instruct &  & 45.69 & 41.41 & 54.91 & 33.12 & 56.51 & 47.01 & 46.82 \\
        Qwen3.5-4B &  & 48.05 & 40.80 & 60.43 & 43.66 & 54.14 & 45.87 & 49.82 \\
        Qwen3-VL-4B-Instruct & 3 & 51.58 & 46.21 & 62.42 & 47.74 & 61.83 & 49.29 & 48.41 \\
        Qwen3.5-9B & 2 & 52.57 & 45.86 & 64.88 & 50.54 & 58.88 & 51.00 & 50.88 \\
        Qwen3.5-4B\textsuperscript{\(\dagger\)} & 1 & 62.89 & 55.54 & 76.07 & 58.28 & 72.49 & 58.97 & 63.07 \\
        \midrule
        \rowcolor{sectionblue}
        \multicolumn{9}{l}{\textbf{\textit{Open-Source Spatial Intelligence Models (3--5B)}}} \\
        VG-LLM-4B &  & 12.59 & 5.41 & 8.28 & 30.54 & 26.04 & 3.70 & 14.84 \\
        Spatial-MLLM-4B & 2 & \secondscore{54.67} & 45.16 & \secondscore{70.86} & 52.90 & 66.57 & \secondscore{51.28} & \secondscore{51.77} \\
        SpatialStack-5B &  & 40.69 & 39.67 & 38.34 & 35.27 & 45.56 & 45.30 & 44.17 \\
        SpatialStack-5B\textsuperscript{\(\dagger\)} & 3 & 53.77 & \secondscore{49.78} & 59.97 & \secondscore{53.12} & \secondscore{67.16} & 49.86 & 49.65 \\
        \textbf{SpatialMind-5B (ours)} & 1 & \bestscore{62.21} & \bestscore{52.05} & \bestscore{75.31} & \bestscore{57.85} & \bestscore{73.67} & \bestscore{62.11} & \bestscore{60.78} \\
        \midrule
        \rowcolor{sectionblue}
        \multicolumn{9}{l}{\textbf{\textit{Open-Source Spatial Intelligence Models (7--10B)}}} \\
        SpatialStack-10B & 3 & 44.81 & 45.42 & 41.56 & 35.05 & 55.62 & 47.58 & 47.17 \\
        GeoThinker (Qwen2.5-VL-7B) &  & 32.14 & 32.96 & 25.15 & 29.89 & 25.15 & 48.43 & 34.45 \\
        GeoThinker (Qwen3-VL-8B) &  & 32.91 & 32.78 & 30.06 & 31.18 & 21.89 & 47.01 & 35.69 \\
        SpatialStack-10B\textsuperscript{\(\dagger\)} & 2 & \secondscore{63.48} & \secondscore{55.80} & \secondscore{75.61} & \bestscore{61.29} & \bestscore{71.01} & \secondscore{61.82} & \bestscore{63.43} \\
        \textbf{SpatialMind-10B (ours)} & 1 & \bestscore{64.54} & \bestscore{57.80} & \bestscore{77.76} & \secondscore{59.57} & \secondscore{70.71} & \bestscore{64.67} & \secondscore{63.25} \\
        \bottomrule
    \end{tabular}
    }%
    \label{tab:sqa3d}
    \endgroup
\end{table*}

\noindent\textbf{SQA3D.}
SQA3D contains 33.4K questions associated with 6.8K
situated contexts from 650 ScanNet scenes. Given a textual description of an
agent's position and orientation, the model must reason about its surrounding
3D environment to answer questions involving spatial relations, navigation,
and multi-hop reasoning. We report top-1 exact-match accuracy using the same
evaluation protocol for all compared models. Detailed results by question
type are reported in Table~\ref{tab:sqa3d}.

\noindent\textbf{BLINK.}
BLINK reformulates 14 classical computer-vision tasks as
3,807 multiple-choice questions with single- or multi-image inputs. It
evaluates fine-grained visual perception abilities, including relative depth
estimation, visual correspondence, and multi-view reasoning. We follow the
official protocol and report the average classification accuracy across all
tasks. Detailed results across the 14 BLINK tasks are reported in
Table~\ref{tab:blink}.

\noindent\textbf{SPAR-Bench.}
SPAR-Bench contains 7,207 manually verified
questions across 20 spatial tasks. It covers depth estimation, distance
prediction, spatial relations, and spatial imagination using both single-view
and multi-view inputs. Because the benchmark contains both numerical
prediction and multiple-choice questions, we follow its official
task-specific evaluation protocol and report the average score across all
tasks.  Detailed results on SPAR-Bench are reported in
Table~\ref{tab:sparbench}.

\noindent\textbf{SPBench.}
SPBench is constructed from the ScanNet
validation set and contains two complementary subsets: SPBench-SI, with 1,009
single-image questions, and SPBench-MV, with 319 multi-view questions. It
evaluates absolute distance, object size, relative distance, and relative
direction, while SPBench-MV additionally evaluates cross-view object
counting. We average the numerical-question and multiple-choice-question
scores within each subset and then average the two subset scores to obtain the
reported SPBench result. Detailed results on SPBench-SI and SPBench-MV are
reported in Table~\ref{tab:spbench}.

\noindent\textbf{Overall Metric.}
The overall score reported in the cross-benchmark comparison is the unweighted
mean of six benchmark-level scores: VSI-Bench and the five additional
benchmarks described above. All compared models are evaluated using the same
input construction, prompting format, and official evaluation scripts.

\noindent\textbf{Discussion of Benchmark-Specific Results.}
Although SpatialMind achieves strong overall performance across the evaluated
benchmarks, its two variants do not rank first on every benchmark.
Both SpatialMind-5B and SpatialMind-10B rank second on BLINK, while
SpatialMind-10B also ranks second on SPBench. This behavior is consistent with
the specialization of our framework. SpatialMind primarily targets grounded
spatial reasoning through task-adaptive fusion of multi-level geometric
representations, whereas BLINK additionally includes broad 2D visual-perception
and visual-logical tasks, such as art-style recognition, forensic reasoning,
jigsaw solving, and visual similarity, for which explicit geometric cues
provide limited benefits.

On SPBench, SpatialMind-10B trails the training-matched SpatialStack-10B by
only $0.59$ points overall ($68.34$ vs.\ $68.93$). The gap is concentrated in
SPBench-SI ($60.95$ vs.\ $63.50$), particularly in absolute-distance and
direction estimation. In contrast, SpatialMind-10B performs better on
SPBench-MV ($75.74$ vs.\ $74.36$), indicating that its remaining limitation
lies mainly in fine-grained single-view metric and directional perception
rather than cross-view aggregation. In addition, our training corpus focuses
primarily on spatial reasoning and provides comparatively limited coverage of
general visual-perception tasks. Incorporating more diverse visual supervision,
stronger metric-geometry estimation, and complementary visual-logical reasoning
modules represents a promising direction for further improvement.

\begin{figure}[t]
    \includegraphics[width=\columnwidth]{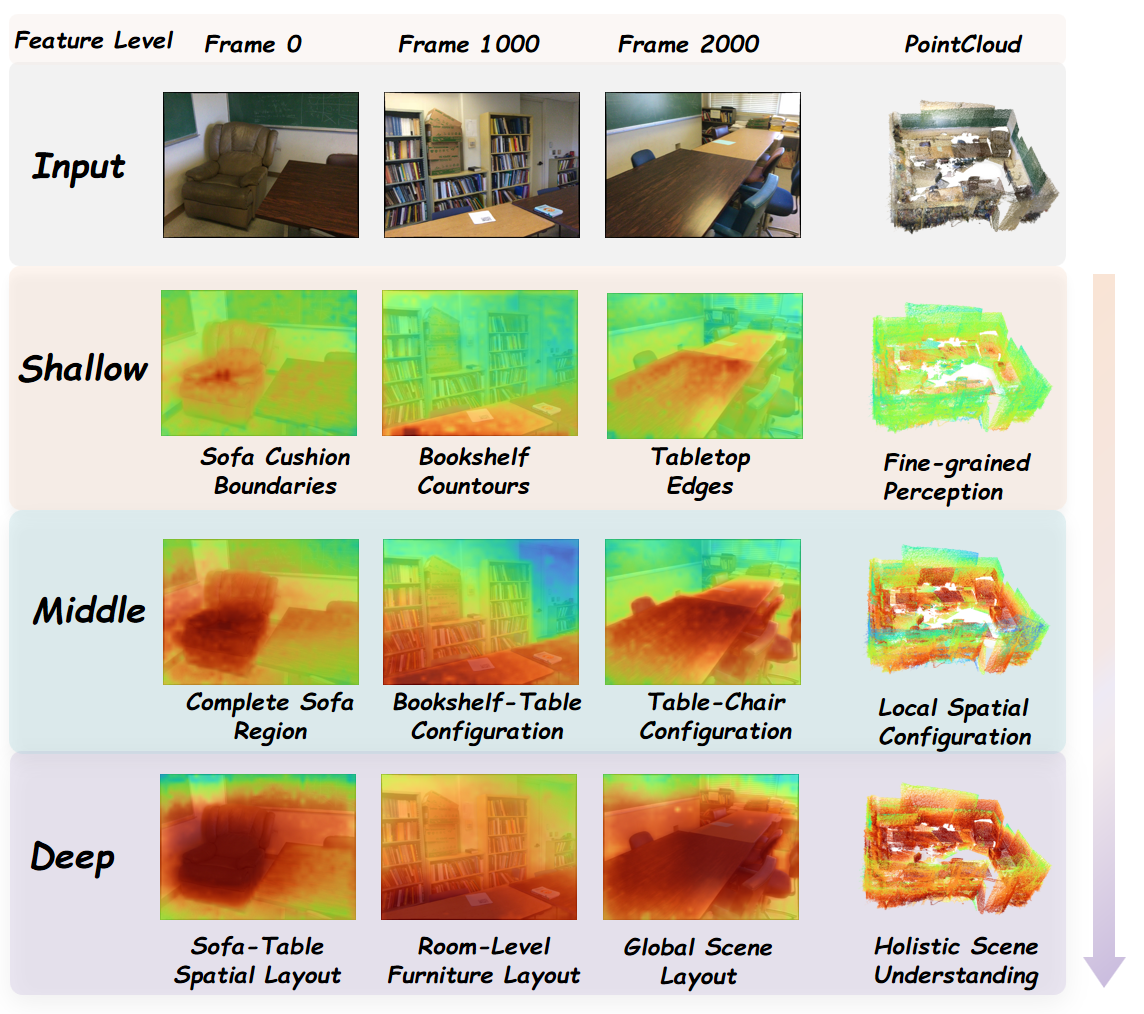}
    \caption{\small
    \textbf{Visualization of Multi-Level Geometric Representations.}
    We visualize similarity maps derived from shallow, intermediate, and deep geometric features across representative video frames and the reconstructed point cloud. }
    \label{fig:Attention Visualization.}
\end{figure}

\section{Complete Results of Multi-Level Geometric Ablation}

\begin{table*}[!t]
    \centering
    \caption{\small{
        \textbf{Evaluation on BLINK.}
        \bestscore{Bold} and \secondscore{underlined} values indicate the
        best and second-best results within each specialized-model size
        group, respectively.
        \textsuperscript{\(\dagger\)} denotes models trained using the same
        training data.
    }}
    \begingroup
    \renewcommand{\arraystretch}{0.95}
    \setlength{\tabcolsep}{2.8pt}
    \setlength{\aboverulesep}{1.4pt}
    \setlength{\belowrulesep}{1.4pt}
    \fontsize{5.3}{6.5}\selectfont

    \newcommand{\rotHead}[1]{%
        \rotatebox[origin=lb]{70}{\textbf{#1}}%
    }

    \resizebox{\textwidth}{!}{%
    \begin{tabular}{@{}
        >{\raggedright\arraybackslash}p{4.0cm}
        *{16}{c}
    @{}}
        \toprule
        \multicolumn{1}{l}{\textbf{Methods}}
        & \rotHead{Rank}
        & \rotHead{Avg.}
        & \rotHead{Art Style}
        & \rotHead{Counting}
        & \rotHead{Forensic}
        & \rotHead{Functional}
        & \rotHead{IQ}
        & \rotHead{Jigsaw}
        & \rotHead{Multi-view}
        & \rotHead{Localization}
        & \rotHead{Rel. Depth}
        & \rotHead{Rel. Reflect.}
        & \rotHead{Semantic}
        & \rotHead{Spatial Rel.}
        & \rotHead{Visual Corr.}
        & \rotHead{Visual Sim.} \\
        \midrule
        \rowcolor{sectionblue}
        \multicolumn{17}{l}{\textbf{\textit{Proprietary Models (API)}}} \\
        GPT-5.6-Luna & 1 & 65.60 & 57.26 & 71.67 & 71.21 & 44.62 & 20.00 & 74.67 & 45.11 & 86.89 & 86.29 & 65.67 & 63.31 & 77.62 & 91.28 & 61.48 \\
        Doubao-Seed-2.0-Pro & 2 & 50.60 & 58.97 & 75.83 & 11.36 & 5.38 & 8.67 & 35.33 & 24.81 & 72.95 & 91.13 & 61.94 & 30.22 & 84.62 & 80.81 & 69.63 \\
        Gemini-3.5-Flash & 3 & 33.88 & 81.20 & 82.50 & 44.70 & 0.00 & 1.33 & 50.67 & 48.12 & 13.11 & 0.00 & 0.00 & 0.00 & 92.31 & 9.30 & 62.96 \\
        \midrule
        \rowcolor{sectionblue}
        \multicolumn{17}{l}{\textbf{\textit{Open-Source Models}}} \\
        Qwen2.5-VL-3B-Instruct & 3 & 48.87 & 63.25 & 65.00 & 28.79 & 21.54 & 22.67 & 49.33 & 44.36 & 47.54 & 74.19 & 42.54 & 35.97 & 82.52 & 45.35 & 67.41 \\
        Qwen2.5-VL-7B-Instruct & 2 & 53.66 & 70.09 & 69.17 & 46.97 & 24.62 & 1.33 & 60.67 & 42.86 & 50.00 & 77.42 & 43.28 & 38.85 & 90.21 & 55.23 & 87.41 \\
        Qwen3.5-4B &  & 29.51 & 71.79 & 73.33 & 59.85 & 0.00 & 0.00 & 38.00 & 29.32 & 0.82 & 0.00 & 0.00 & 0.72 & 82.52 & 1.16 & 68.15 \\
        Qwen3-VL-4B-Instruct &  & 47.40 & 73.50 & 60.00 & 81.82 & 2.31 & 8.67 & 73.33 & 36.84 & 1.64 & 25.81 & 47.76 & 4.32 & 82.52 & 79.07 & 75.56 \\
        Qwen3.5-9B &  & 42.45 & 56.41 & 49.17 & 71.21 & 0.00 & 0.67 & 68.00 & 37.59 & 22.95 & 85.48 & 44.78 & 4.32 & 87.41 & 13.95 & 63.70 \\
        Qwen3.5-4B\textsuperscript{\(\dagger\)} & 1 & 57.92 & 72.65 & 65.83 & 54.55 & 49.23 & 24.00 & 52.67 & 27.82 & 48.36 & 75.00 & 38.06 & 56.83 & 74.83 & 91.28 & 76.30 \\
        \midrule
        \rowcolor{sectionblue}
        \multicolumn{17}{l}{\textbf{\textit{Open-Source Spatial Intelligence Models (3--5B)}}} \\
        VG-LLM-4B &  & 38.19 & 47.01 & 42.50 & 25.00 & 21.54 & 20.00 & 52.67 & \secondscore{44.36} & 43.44 & 51.61 & 26.87 & 32.37 & 54.55 & 29.65 & 47.41 \\
        Spatial-MLLM-4B &  & 41.45 & 48.72 & \secondscore{57.50} & 26.52 & 21.54 & 25.33 & 52.67 & \bestscore{48.87} & 48.36 & 63.71 & 29.85 & 31.65 & \secondscore{67.13} & 20.35 & 47.41 \\
        SpatialStack-5B & 1 & \bestscore{59.13} & \bestscore{76.07} & \bestscore{61.67} & 47.73 & \bestscore{44.62} & \bestscore{28.00} & \bestscore{66.00} & 14.29 & \bestscore{55.74} & \secondscore{69.35} & \bestscore{47.76} & \bestscore{64.03} & \bestscore{67.83} & \bestscore{93.02} & \bestscore{85.93} \\
        SpatialStack-5B\textsuperscript{\(\dagger\)} & 3 & 53.97 & \secondscore{70.94} & 52.50 & \bestscore{62.88} & 30.77 & 24.00 & 63.33 & 10.53 & 48.36 & \bestscore{79.03} & 37.31 & 45.32 & 60.84 & \secondscore{91.28} & 72.59 \\
        \textbf{SpatialMind-5B (ours)} & 2 & \secondscore{54.36} & \secondscore{70.94} & 52.50 & \secondscore{59.09} & \secondscore{31.54} & \secondscore{27.33} & \secondscore{65.33} & 7.52 & \secondscore{50.00} & \bestscore{79.03} & \secondscore{39.55} & \secondscore{46.04} & 62.24 & \secondscore{91.28} & \secondscore{77.78} \\
        \midrule
        \rowcolor{sectionblue}
        \multicolumn{17}{l}{\textbf{\textit{Open-Source Spatial Intelligence Models (7--10B)}}} \\
        SpatialStack-10B & 1 & \bestscore{64.28} & \bestscore{76.92} & 60.83 & \bestscore{81.06} & \bestscore{53.85} & \bestscore{28.67} & \bestscore{82.00} & 5.26 & \bestscore{55.74} & \secondscore{83.87} & \secondscore{45.52} & \bestscore{60.43} & \bestscore{81.12} & \bestscore{93.02} & \bestscore{85.93} \\
        GeoThinker (Qwen2.5-VL-7B) &  & 48.87 & 69.23 & \secondscore{63.33} & 34.09 & 24.62 & 24.67 & 47.33 & \bestscore{19.55} & \secondscore{54.92} & 83.06 & 38.81 & 30.94 & 61.54 & 61.05 & 76.30 \\
        GeoThinker (Qwen3-VL-8B) &  & 53.92 & 68.38 & 60.00 & \secondscore{71.97} & 25.38 & 23.33 & 60.67 & \secondscore{18.80} & 48.36 & \bestscore{93.55} & 42.54 & 32.37 & 63.64 & 67.44 & 81.48 \\
        SpatialStack-10B\textsuperscript{\(\dagger\)} & 3 & 56.13 & 71.79 & 60.00 & 41.67 & 40.77 & \secondscore{27.33} & 65.33 & 9.02 & 42.62 & 79.84 & 37.31 & \secondscore{56.83} & 72.73 & \secondscore{92.44} & 80.74 \\
        \textbf{SpatialMind-10B (ours)} & 2 & \secondscore{60.02} & \secondscore{76.07} & \bestscore{69.17} & 65.15 & \secondscore{43.08} & 25.33 & \secondscore{72.67} & 4.51 & 51.64 & 75.00 & \bestscore{50.75} & 50.36 & \secondscore{77.62} & 91.86 & \secondscore{82.22} \\
        \bottomrule
    \end{tabular}
    }%
    \label{tab:blink}
    \endgroup
\end{table*}

\begin{table}[t]
    \centering
    \caption{\textbf{Ablation results on low-level spatial tasks (\%).}}
    \label{tab:low_level_ablation}
    
    \scriptsize
    \setlength{\tabcolsep}{2.0pt}
    \renewcommand{\arraystretch}{1.08}
    
    \resizebox{\columnwidth}{!}{%
    \begin{tabular}{@{}lccccccc@{}}
        \toprule
        \textbf{Model}
        & \makecell{\textbf{Abs.}\\\textbf{Distance}}
        & \makecell{\textbf{Rel.}\\\textbf{Distance}}
        & \makecell{\textbf{Object}\\\textbf{Size}}
        & \makecell{\textbf{Room}\\\textbf{Size}}
        & \makecell{\textbf{Rel.}\\\textbf{Direction}}
        & \textbf{Avg.}
        & \makecell{\textbf{Change}\\\textbf{(pp)}} \\
        \midrule
        
        Full model
        & 55.70
        & 66.76
        & 69.18
        & 68.09
        & 84.51
        & 68.85
        & -- \\
        
        w/o 11--0
        & 51.07
        & 66.90
        & 69.67
        & 67.77
        & 81.71
        & 67.42
        & $-1.43$ \\
        
        w/o 17--1
        & 53.50
        & 65.77
        & 67.36
        & 67.36
        & 83.92
        & 67.52
        & $-1.33$ \\
        
        w/o 23--2
        & 54.19
        & 65.21
        & 69.41
        & 69.06
        & 83.12
        & 68.19
        & $-0.66$ \\
        
        \bottomrule
    \end{tabular}
    }
\end{table}

\begin{table}[t]
    \centering
    \caption{\textbf{Ablation results on high-level spatial tasks (\%).}}
    \label{tab:high_level_ablation}
    
    \scriptsize
    \setlength{\tabcolsep}{2.2pt}
    \renewcommand{\arraystretch}{1.08}
    
    \begin{tabularx}{\columnwidth}{@{}
        >{\raggedright\arraybackslash}p{1.15cm}
        *{5}{>{\centering\arraybackslash}X}
    @{}}
        \toprule
        \textbf{Model}
        & \makecell{\textbf{Appearance}\\\textbf{Order}}
        & \makecell{\textbf{Object}\\\textbf{Counting}}
        & \makecell{\textbf{Route}\\\textbf{Planning}}
        & \textbf{Avg.}
        & \makecell{\textbf{Change}\\\textbf{(pp)}} \\
        \midrule
        
        Full model
        & 83.33
        & 70.86
        & 42.26
        & 65.48
        & -- \\
        
        w/o 11--0
        & 83.17
        & 69.98
        & 42.78
        & 65.31
        & $-0.17$ \\
        
        w/o 17--1
        & 81.55
        & 69.43
        & 40.72
        & 63.90
        & $-1.58$ \\
        
        w/o 23--2
        & 81.22
        & 70.88
        & 38.14
        & 63.41
        & $-2.07$ \\
        
        \bottomrule
    \end{tabularx}
\end{table}

\textbf{Visualization Method.}
To examine the hierarchical differences among multi-level geometric
representations, we visualize the features extracted from the shallow,
intermediate, and deep layers of the geometry encoder. Specifically, we first
reshape the geometric tokens at each feature level back into their original
spatial layout. We then select a small patch as the region of interest (ROI)
and average its feature tokens to obtain an ROI representation. For each
spatial location, we compute the cosine similarity between its feature vector
and the ROI representation, producing a patch-wise similarity map. The
similarity scores are normalized and overlaid on the corresponding image or
point-cloud view, where brighter regions indicate higher similarity and
darker regions indicate lower similarity.

\textbf{Qualitative Results.}
As shown in Fig.~\ref{fig:Attention Visualization.}, geometric features
exhibit clear hierarchical differences. Shallow features preserve
fine-grained spatial details and respond strongly to local geometric
structures, such as sofa cushion boundaries, bookshelf contours, tabletop
edges, and detailed point-cloud surfaces. Intermediate features expand their
responses from isolated boundaries to more complete object regions and local
spatial configurations, including the complete sofa region, bookshelf--table
configuration, and table--chair arrangement. In contrast, deep features
produce broader and more spatially coherent responses, emphasizing
object-level relationships, room-level furniture layouts, global scene
structures, and the holistic organization of the reconstructed point cloud.
These visualization results suggest that shallow representations primarily
encode local geometric details, intermediate representations capture
object-level configurations, and deep representations progressively aggregate
global scene context.

\textbf{Quantitative Results.}
The task-level ablation results in
Tables~\ref{tab:low_level_ablation} and~\ref{tab:high_level_ablation}
quantitatively support the hierarchical patterns observed in
Fig.~\ref{fig:Attention Visualization.}. The complete multi-level
configuration achieves average scores of $68.85\%$ and $65.48\%$ on
low-level perception and high-level reasoning tasks, respectively,
outperforming every single-layer removal variant.

As shown in Table~\ref{tab:low_level_ablation}, removing the shallow
injection (VGGT-11 $\rightarrow$ LLM-0) causes the largest degradation on
low-level tasks, reducing the average score from $68.85\%$ to $67.42\%$, a
drop of $1.43$ percentage points. Its effect on high-level tasks is
comparatively small, decreasing the average by only $0.17$ points. This
behavior is consistent with the visualization results, where shallow features
highlight object boundaries, local depth variations, and fine-grained
geometric details that are particularly important for distance, size, and
direction estimation.

In contrast, Table~\ref{tab:high_level_ablation} shows that removing the deep
injection (VGGT-23 $\rightarrow$ LLM-2) produces the largest degradation on
high-level tasks, reducing the average score from $65.48\%$ to $63.41\%$, a
drop of $2.07$ percentage points. The corresponding reduction on low-level
tasks is only $0.66$ points. This result agrees with the deep-feature
visualizations, which emphasize room-level layouts, cross-object relations,
and holistic scene structures required by route planning, object counting,
and appearance-order reasoning.

Removing the intermediate injection
(VGGT-17 $\rightarrow$ LLM-1) decreases the low- and high-level averages by
$1.33$ and $1.58$ percentage points, respectively. This more balanced
degradation is consistent with the intermediate similarity maps, which
capture complete object regions and local object configurations, thereby
bridging fine-grained geometric perception and global scene reasoning.

\begin{table*}[!t]
    \centering
    \caption{\small{
        \textbf{Evaluation on SPAR-Bench.}
        \bestscore{Bold} and \secondscore{underlined} values indicate the
        best and second-best results within each specialized-model size
        group, respectively.
        \textsuperscript{\(\dagger\)} denotes models trained using the same
        training data.
    }}
    \label{tab:sparbench}

    \begingroup
    \renewcommand{\arraystretch}{0.95}
    \setlength{\tabcolsep}{2.8pt}
    \setlength{\aboverulesep}{1.4pt}
    \setlength{\belowrulesep}{1.4pt}
    \fontsize{5.3}{6.8}\selectfont

    \newcommand{\rotHead}[1]{%
        \rotatebox[origin=lb]{70}{\textbf{#1}}%
    }

    \newcommand{\rotGroup}[1]{%
        \rotatebox[origin=lb]{70}{\textbf{\textit{#1}}}%
    }

    \resizebox{\textwidth}{!}{%
    \begin{tabular}{@{}
        >{\raggedright\arraybackslash}p{3.6cm}
        cc
        !{\vrule width 0.35pt}
        c*{8}{c}
        !{\vrule width 0.35pt}
        c*{3}{c}
        !{\vrule width 0.35pt}
        c*{9}{c}
        @{}}

        \toprule
        \multicolumn{1}{l}{\textbf{Methods}}
        & \rotHead{Rank}
        & \rotHead{Avg.}

        & \rotGroup{Low}
        & \rotHead{Depth-OC}
        & \rotHead{Depth-OC-MV}
        & \rotHead{Depth-OO}
        & \rotHead{Depth-OO-MV}
        & \rotHead{Dist-OC}
        & \rotHead{Dist-OC-MV}
        & \rotHead{Dist-OO}
        & \rotHead{Dist-OO-MV}

        & \rotGroup{Medium}
        & \rotHead{PosMatch}
        & \rotHead{CamMotion}
        & \rotHead{ViewChg}

        & \rotGroup{High}
        & \rotHead{Dist-C-OO}
        & \rotHead{Dist-C-OO-MV}
        & \rotHead{ObjRel-OC-MV}
        & \rotHead{ObjRel-OO}
        & \rotHead{ObjRel-OO-MV}
        & \rotHead{SphImag-OC}
        & \rotHead{SphImag-OC-MV}
        & \rotHead{SphImag-OO}
        & \rotHead{SphImag-OO-MV} \\
        \midrule

        \rowcolor{sectionblue}
        \multicolumn{26}{l}{
            \textbf{\textit{Proprietary Models (API)}}
        } \\

        GPT-5.6-Luna
        & 3 & 47.65
        & 46.51
        & 60.22 & 48.91 & 27.98 & 26.32
        & 58.65 & 59.32 & 51.10 & 39.60
        & 41.17
        & 74.55 & 22.75 & 26.21
        & 50.82
        & 71.47 & 61.90
        & 46.50 & 66.76 & 41.83
        & 45.16 & 42.44 & 40.73 & 40.62 \\

        Doubao-Seed-2.0-Pro
        & 2 & 55.50
        & 46.85
        & 74.22 & 57.84 & 31.61 & 24.41
        & 69.13 & 63.81 & 29.70 & 24.11
        & 50.26
        & 86.26 & 35.25 & 29.28
        & 64.92
        & 82.94 & 80.65
        & 62.00 & 79.12 & 69.53
        & 53.76 & 51.45 & 47.68 & 57.14 \\

        Gemini-3.5-Flash
        & 1 & 57.23
        & 55.98
        & 75.25 & 63.19 & 34.22 & 32.85
        & 74.66 & 62.38 & 63.28 & 42.02
        & 32.10
        & 45.55 & 25.25 & 25.49
        & 66.72
        & 87.65 & 76.19
        & 69.25 & 77.20 & 73.41
        & 62.90 & 54.94 & 53.31 & 45.66 \\

        \midrule

        \rowcolor{sectionblue}
        \multicolumn{26}{l}{
            \textbf{\textit{Open-Source Models}}
        } \\

        Qwen2.5-VL-3B-Instruct
        &  & 29.53
        & 25.66
        & 38.58 & 40.53 & 18.25 & 17.72
        & 16.62 & 21.32 & 29.97 & 22.26
        & 28.15
        & 43.77 & 27.00 & 13.67
        & 33.43
        & 57.06 & 58.63
        & 39.75 & 30.49 & 20.78
        & 19.89 & 26.74 & 22.85 & 24.65 \\

        Qwen2.5-VL-7B-Instruct
        &  & 34.19
        & 27.06
        & 32.50 & 34.62 & 21.08 & 10.94
        & 40.87 & 29.21 & 23.69 & 23.60
        & 30.39
        & 45.04 & 26.50 & 19.64
        & 41.79
        & 58.53 & 55.95
        & 43.50 & 51.10 & 38.23
        & 33.06 & 33.43 & 29.80 & 32.49 \\

        Qwen3.5-4B
        & 3 & 39.84
        & 41.22
        & 56.69 & 48.91 & 22.72 & 15.99
        & 57.05 & 47.73 & 43.03 & 37.61
        & 35.71
        & 60.56 & 25.00 & 21.58
        & 39.99
        & 69.41 & 61.31
        & 50.00 & 47.25 & 31.86
        & 24.46 & 27.62 & 24.17 & 23.81 \\

        Qwen3-VL-4B-Instruct
        &  & 36.25
        & 29.92
        & 33.89 & 38.09 & 18.20 & 19.27
        & 14.17 & 20.36 & 51.05 & 44.31
        & 33.58
        & 52.67 & 29.75 & 18.33
        & 42.78
        & 67.65 & 61.01
        & 44.75 & 53.85 & 33.24
        & 27.42 & 36.92 & 30.46 & 29.69 \\

        Qwen3.5-9B
        & 2 & 46.69
        & 45.98
        & 61.67 & 51.97 & 25.05 & 22.18
        & 57.10 & 53.73 & 50.52 & 45.62
        & 38.52
        & 61.32 & 32.50 & 21.73
        & 50.05
        & 68.24 & 67.56
        & 55.25 & 62.91 & 41.27
        & 37.90 & 40.70 & 33.77 & 42.86 \\

        Qwen3.5-4B\textsuperscript{\(\dagger\)}
        & 1 & 68.54
        & 59.96
        & 82.69 & 78.56 & 30.32 & 22.77
        & 77.56 & 76.63 & 61.90 & 49.23
        & 61.28
        & 87.02 & 65.00 & 31.81
        & 78.59
        & 87.65 & 81.25
        & 87.75 & 84.89 & 82.55
        & 72.58 & 71.80 & 63.25 & 75.63 \\

        \midrule

        \rowcolor{sectionblue}
        \multicolumn{26}{l}{
            \textbf{\textit{
                Open-Source Spatial Intelligence Models (3--5B)
            }}
        } \\

        VG-LLM-4B
        &  & 20.66
        & 12.92
        & 10.81 & 8.53 & 20.11 & 19.01
        & 5.37 & 4.74 & 17.77 & 17.07
        & 22.97
        & 20.36 & 23.50 & 25.05
        & 26.76
        & 46.18 & 43.45
        & 24.25 & 20.05 & 20.22
        & 21.77 & 21.22 & 19.87 & 23.81 \\

        Spatial-MLLM-4B
        &  & 32.51
        & 30.07
        & 32.97 & 29.41 & 18.23 & 20.43
        & 31.25 & 35.56 & 36.72 & 36.03
        & 26.75
        & 37.15 & 35.75 & 7.35
        & 36.60
        & 54.41 & 55.65
        & 41.25 & 36.81 & 21.33
        & 30.65 & 30.81 & 30.79 & 27.73 \\

        SpatialStack-5B
        & 3 & 68.33
        & \bestscore{57.33}
        & \bestscore{80.69}
        & \secondscore{74.94}
        & 23.55
        & 17.63
        & \bestscore{74.63}
        & \bestscore{74.96}
        & \bestscore{64.77}
        & \bestscore{47.47}
        & 63.54
        & \bestscore{87.79}
        & 67.25
        & \secondscore{35.58}
        & 79.71
        & \bestscore{90.59}
        & \bestscore{86.01}
        & 89.25
        & 83.79
        & 83.10
        & 73.66
        & 72.97
        & 64.90
        & 73.11 \\

        SpatialStack-5B\textsuperscript{\(\dagger\)}
        & 2
        & \secondscore{69.70}
        & 56.11
        & 78.17
        & 74.22
        & \bestscore{25.67}
        & \bestscore{22.53}
        & 71.15
        & 73.29
        & 59.61
        & 44.21
        & \secondscore{64.08}
        & 85.50
        & \bestscore{72.25}
        & 34.48
        & \secondscore{83.65}
        & \secondscore{89.12}
        & \secondscore{85.71}
        & \bestscore{91.00}
        & \bestscore{87.36}
        & \bestscore{87.26}
        & \secondscore{80.91}
        & \bestscore{77.91}
        & \secondscore{76.82}
        & \secondscore{76.75} \\

        \textbf{SpatialMind-5B (ours)}
        & 1
        & \bestscore{70.00}
        & \secondscore{56.68}
        & \secondscore{79.36}
        & \bestscore{75.44}
        & \secondscore{25.35}
        & \secondscore{20.70}
        & \secondscore{71.86}
        & \secondscore{74.88}
        & \secondscore{60.33}
        & \secondscore{45.49}
        & \bestscore{65.14}
        & \secondscore{86.01}
        & \secondscore{72.00}
        & \bestscore{37.42}
        & \bestscore{83.76}
        & 88.53
        & \secondscore{85.71}
        & \secondscore{90.75}
        & \secondscore{86.81}
        & \secondscore{86.43}
        & \bestscore{81.72}
        & \secondscore{77.62}
        & \bestscore{78.15}
        & \bestscore{78.15} \\

        \midrule

        \rowcolor{sectionblue}
        \multicolumn{26}{l}{
            \textbf{\textit{
                Open-Source Spatial Intelligence Models (7--10B)
            }}
        } \\

        SpatialStack-10B
        & 2
        & \secondscore{69.66}
        & 59.56
        & \secondscore{82.94}
        & 76.50
        & 28.84
        & 21.13
        & \secondscore{76.44}
        & \secondscore{77.42}
        & \secondscore{64.24}
        & 48.92
        & \bestscore{64.28}
        & \bestscore{87.53}
        & 66.75
        & \bestscore{38.55}
        & 80.44
        & \secondscore{88.82}
        & 83.33
        & 89.25
        & \secondscore{86.26}
        & 84.76
        & 71.24
        & \secondscore{73.84}
        & \secondscore{70.53}
        & 75.91 \\

        GeoThinker (Qwen2.5-VL-7B)
        &  & 68.38
        & 58.47
        & 79.72
        & 74.50
        & 23.12
        & 20.94
        & 75.62
        & 77.04
        & 61.07
        & \bestscore{55.76}
        & 58.34
        & 75.83
        & \secondscore{67.00}
        & 32.20
        & \secondscore{80.54}
        & 86.18
        & 83.93
        & \bestscore{90.75}
        & 81.87
        & \bestscore{86.98}
        & \bestscore{79.03}
        & \bestscore{79.07}
        & 62.25
        & 74.79 \\

        GeoThinker (Qwen3-VL-8B)
        & 3
        & 69.12
        & \bestscore{61.54}
        & \bestscore{84.17}
        & \bestscore{77.53}
        & \secondscore{31.48}
        & 21.29
        & \bestscore{80.56}
        & \bestscore{79.32}
        & 63.11
        & \secondscore{54.85}
        & 60.09
        & 80.92
        & \secondscore{67.00}
        & 32.36
        & 78.88
        & 87.94
        & 77.38
        & 87.50
        & 85.16
        & 84.76
        & 75.81
        & 70.64
        & 65.89
        & 74.79 \\

        SpatialStack-10B\textsuperscript{\(\dagger\)}
        &  & 68.80
        & 58.42
        & 82.72
        & \secondscore{76.75}
        & 27.26
        & \secondscore{23.52}
        & 74.12
        & 74.52
        & 62.48
        & 45.96
        & \secondscore{62.12}
        & \secondscore{82.44}
        & \bestscore{68.75}
        & 35.16
        & 80.25
        & 87.35
        & \secondscore{84.52}
        & 88.75
        & \bestscore{87.09}
        & 85.04
        & \secondscore{76.88}
        & 71.51
        & 63.25
        & \secondscore{77.87} \\

        \textbf{SpatialMind-10B (ours)}
        & 1
        & \bestscore{70.27}
        & \secondscore{59.75}
        & 80.42
        & 76.16
        & \bestscore{35.81}
        & \bestscore{24.57}
        & 72.80
        & 75.75
        & \bestscore{65.84}
        & 46.67
        & 61.18
        & 80.92
        & 64.75
        & \secondscore{37.86}
        & \bestscore{82.65}
        & \bestscore{90.59}
        & \bestscore{88.99}
        & \secondscore{89.50}
        & 85.44
        & \secondscore{85.87}
        & 75.27
        & \secondscore{73.84}
        & \bestscore{70.86}
        & \bestscore{83.47} \\

        \bottomrule
    \end{tabular}
    }%

    \endgroup
\end{table*}

\noindent\textbf{Overall Analysis.}
The qualitative and quantitative results jointly reveal relative hierarchical
specialization among multi-level geometric representations. Shallow features
contribute more strongly to fine-grained spatial perception, intermediate
features provide object-level and relational cues across both task
hierarchies, and deep features are more important for global spatial
reasoning. Meanwhile, the complete configuration consistently outperforms all
single-layer removal variants on both task groups, demonstrating that these
feature levels are not redundant but complementary. Their joint use enables
the model to preserve local geometric details while progressively integrating
object-level configurations and holistic scene structures.

\begin{table*}[!t]
    \centering
    \caption{\small{\textbf{Evaluation on SPBench.} \bestscore{Bold} and \secondscore{underlined} values indicate the best and second-best results within each specialized-model size group, respectively. \textsuperscript{\(\dagger\)} denotes models trained using the same training data.}}
    \begingroup
    \renewcommand{\arraystretch}{0.95}
    \setlength{\tabcolsep}{2.8pt}
    \setlength{\aboverulesep}{1.4pt}
    \setlength{\belowrulesep}{1.4pt}
    \fontsize{5.3}{6.5}\selectfont
    \resizebox{\textwidth}{!}{%
    \begin{tabular}{>{\raggedright\arraybackslash}p{4.8cm}ccccccccccccc}
        \toprule
        \multirow{2}{*}{\textbf{Methods}} & \multirow{2}{*}{\textbf{Rank}} & \multicolumn{5}{c}{\cellcolor{headercream}\textbf{SPBench-SI}} & \multicolumn{6}{c}{\cellcolor{headercream}\textbf{SPBench-MV}} & \multirow{2}{*}{\textbf{Overall}} \\
        \cmidrule(lr){3-7} \cmidrule(lr){8-13}
        & & Overall & Abs. Dist. & Direction & Rel. Dist. & Size & Overall & Abs. Dist. & Counting & Direction & Rel. Dist. & Size & Mean \\
        \midrule
        \rowcolor{sectionblue}
        \multicolumn{14}{l}{\textbf{\textit{Proprietary Models (API)}}} \\
        GPT-5.6-Luna & 2 & 54.72 & 41.01 & 33.66 & 79.12 & 65.08 & 53.66 & 32.33 & 50.86 & 29.55 & 88.24 & 67.34 & 54.19 \\
        Doubao-Seed-2.0-Pro & 3 & 48.06 & 2.55 & 31.70 & 90.11 & 67.88 & 52.49 & 19.67 & 55.57 & 25.00 & 94.12 & 68.10 & 50.28 \\
        Gemini-3.5-Flash & 1 & 58.19 & 33.62 & 30.07 & 92.31 & 76.76 & 65.34 & 42.33 & 56.57 & 56.82 & 94.12 & 76.84 & 61.77 \\
        \midrule
        \rowcolor{sectionblue}
        \multicolumn{14}{l}{\textbf{\textit{Open-Source Models}}} \\
        Qwen2.5-VL-3B-Instruct &  & 40.95 & 24.70 & 34.97 & 76.92 & 27.19 & 40.01 & 29.00 & 48.86 & 15.91 & 82.35 & 23.92 & 40.48 \\
        Qwen2.5-VL-7B-Instruct &  & 47.05 & 21.34 & 41.18 & 78.02 & 47.65 & 39.45 & 8.33 & 47.43 & 11.36 & 94.12 & 36.01 & 43.25 \\
        Qwen3.5-4B &  & 50.60 & 14.77 & 34.97 & 91.21 & 61.47 & 55.80 & 40.67 & 52.57 & 22.73 & 100.00 & 63.04 & 53.20 \\
        Qwen3-VL-4B-Instruct & 2 & 50.35 & 14.70 & 29.74 & 85.71 & 71.23 & 64.73 & 45.00 & 76.00 & 31.82 & 94.12 & 76.71 & 57.54 \\
        Qwen3.5-9B & 3 & 48.93 & 10.13 & 32.03 & 86.81 & 66.76 & 57.53 & 37.67 & 58.71 & 25.00 & 100.00 & 66.27 & 53.23 \\
        Qwen3.5-4B\textsuperscript{\(\dagger\)} & 1 & 59.01 & 25.30 & 48.37 & 86.81 & 75.57 & 75.97 & 48.67 & 93.14 & 68.18 & 94.12 & 75.76 & 67.49 \\
        \midrule
        \rowcolor{sectionblue}
        \multicolumn{14}{l}{\textbf{\textit{Open-Source Spatial Intelligence Models (3--5B)}}} \\
        VG-LLM-4B &  & 29.84 & 31.81 & 26.14 & 48.35 & 13.07 & 34.58 & 17.33 & 46.14 & 25.00 & 76.47 & 7.97 & 32.21 \\
        Spatial-MLLM-4B &  & 53.34 & 25.77 & 43.79 & 80.22 & 63.59 & 56.09 & 26.00 & \bestscore{88.71} & 22.73 & 76.47 & 66.52 & 54.72 \\
        SpatialStack-5B & 3 & 60.09 & 28.93 & \bestscore{53.59} & 84.62 & \secondscore{73.22} & \secondscore{77.26} & \bestscore{57.33} & 81.00 & 75.00 & \bestscore{100.00} & \secondscore{72.97} & 68.68 \\
        SpatialStack-5B\textsuperscript{\(\dagger\)} & 2 & \secondscore{62.59} & \bestscore{40.63} & \secondscore{51.73} & \secondscore{85.39} & 72.60 & 77.21 & \secondscore{55.64} & 84.42 & \secondscore{78.49} & \secondscore{95.93} & 71.58 & \secondscore{69.90} \\
        \textbf{SpatialMind-5B (ours)} & 1 & \bestscore{64.41} & \secondscore{39.13} & \bestscore{53.59} & \bestscore{89.01} & \bestscore{75.92} & \bestscore{79.45} & 55.33 & \secondscore{85.29} & \bestscore{81.82} & \bestscore{100.00} & \bestscore{74.81} & \bestscore{71.63} \\
        \midrule
        \rowcolor{sectionblue}
        \multicolumn{14}{l}{\textbf{\textit{Open-Source Spatial Intelligence Models (7--10B)}}} \\
        SpatialStack-10B & 3 & 58.23 & 21.95 & 47.06 & 90.11 & 73.82 & \bestscore{78.08} & \bestscore{56.67} & \bestscore{93.14} & \secondscore{65.91} & \bestscore{100.00} & 74.68 & 68.16 \\
        GeoThinker (Qwen2.5-VL-7B) &  & 59.64 & 23.49 & 50.00 & \bestscore{94.51} & 70.58 & 53.63 & 32.33 & 19.14 & 47.73 & \secondscore{94.12} & \secondscore{74.81} & 56.64 \\
        GeoThinker (Qwen3-VL-8B) &  & 60.25 & 26.51 & \secondscore{51.96} & \secondscore{92.31} & 70.24 & 62.05 & 20.33 & 57.86 & \secondscore{65.91} & 88.24 & \bestscore{77.91} & 61.15 \\
        SpatialStack-10B\textsuperscript{\(\dagger\)} & 1 & \bestscore{63.50} & \bestscore{34.43} & \bestscore{55.56} & 87.91 & \secondscore{76.09} & 74.36 & \secondscore{51.67} & 88.14 & 63.64 & \secondscore{94.12} & 74.24 & \bestscore{68.93} \\
        \textbf{SpatialMind-10B (ours)} & 2 & \secondscore{60.95} & \secondscore{30.13} & 49.35 & 86.81 & \bestscore{77.52} & \secondscore{75.74} & 44.00 & \secondscore{91.00} & \bestscore{70.45} & \bestscore{100.00} & 73.23 & \secondscore{68.34} \\
        \bottomrule
    \end{tabular}
    }%
    \label{tab:spbench}
    \endgroup
\end{table*}

\begin{figure}[t]
    \includegraphics[width=\columnwidth]{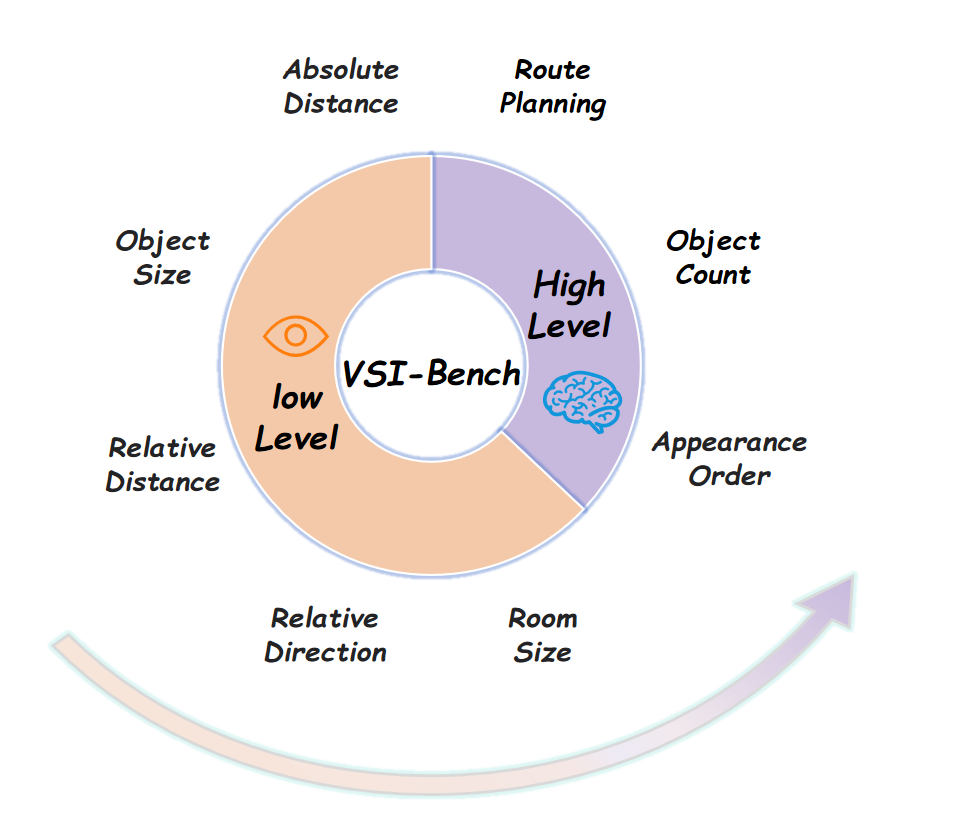}
    \caption{\small
    \textbf{VSI-Bench Task Taxonomy.}
    The eight VSI-Bench subtasks are grouped into low-level perception and
high-level reasoning according to their dominant spatial operations}
    \label{fig:vsi_task_taxonomy}
\end{figure}

\section{Additional Analysis and Visualization of GeoFus}
\subsection{VSI-Bench Task Taxonomy}
As illustrated in Fig.~\ref{fig:vsi_task_taxonomy}, we follow the general
hierarchy of spatial capabilities in SPAR~\cite{zhang2026flatland} and divide
the eight VSI-Bench subtasks into low-level perception and high-level
reasoning according to the dominant operation required to obtain the answer.
The low-level group includes \textit{absolute distance}, \textit{relative
distance}, \textit{relative direction}, \textit{object size}, and
\textit{room size}. These tasks primarily require the direct extraction of
metric, directional, or scale-related cues from visual observations.

The high-level group includes \textit{route planning}, \textit{appearance
order}, and \textit{object counting}. These tasks require cross-view
information integration, global scene modeling, or multi-step reasoning over
spatial relations. Although object counting also relies on perceptual
evidence, we categorize it as a high-level task because counting in
VSI-Bench commonly requires cross-view aggregation and global object-identity
consistency rather than isolated object recognition. The upward arrow in
Fig.~\ref{fig:vsi_task_taxonomy} represents the increasing reasoning
complexity from local spatial perception to global spatial reasoning. This
taxonomy is used only for analysis, while all eight subtasks retain their
original evaluation protocols.

\subsection{Per-Task Adaptive Gating Analysis}

\textbf{Visualization Protocol.}
To examine whether the sample-wise gating decisions learned by GeoFus produce
task-dependent geometric selection patterns, we randomly sample 50 examples
from each of the eight VSI-Bench subtasks. For every example, we record the
softmax-normalized weights assigned to the shallow, intermediate, and deep
geometric representations at the three LLM injection positions
$L_0$, $L_1$, and $L_2$. We then average the weights over the sampled
examples within each subtask. For the group-level comparison, we further
compute macro averages over the five low-level perception tasks and the three
high-level reasoning tasks, ensuring that each subtask contributes equally
regardless of its original dataset size.

\begin{figure*}[t]
    \centering
    \includegraphics[width=0.98\textwidth]
    {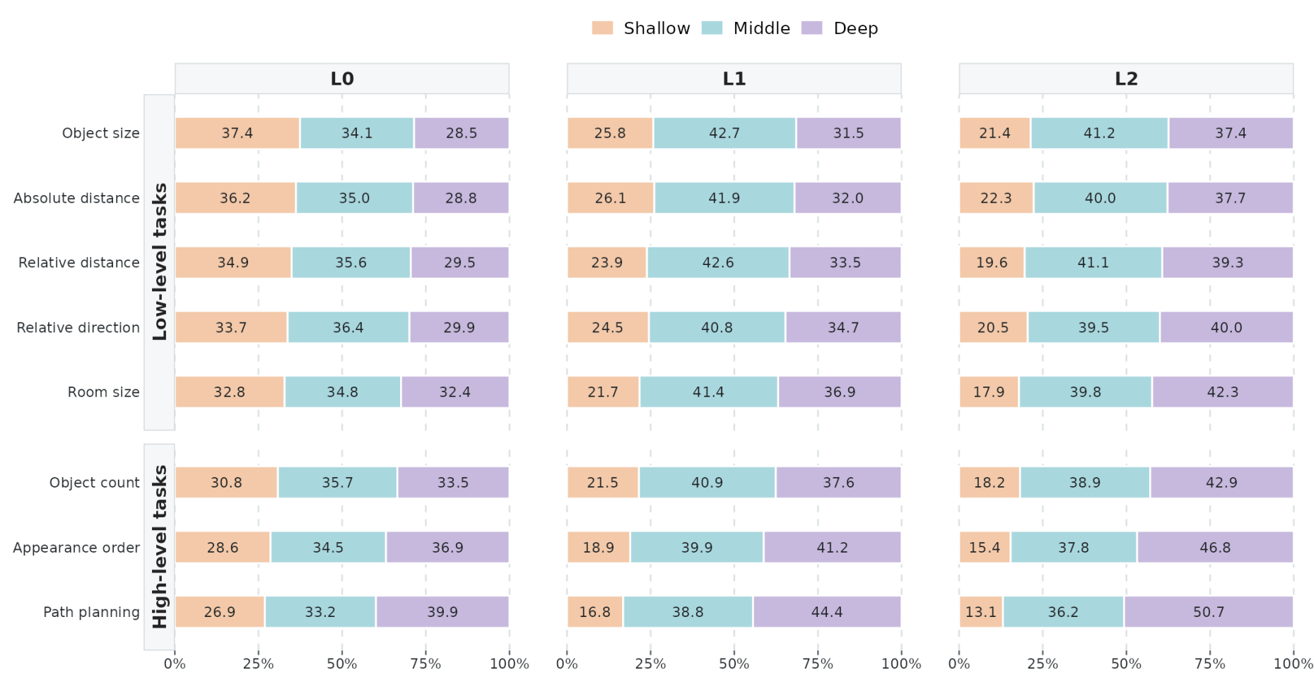}
    \caption{\small{
        \textbf{Per-Task Adaptive Gating Distributions on VSI-Bench.}
        Softmax-normalized gating weights assigned to shallow, intermediate,
        and deep geometric representations at the three LLM injection
        positions, $L_0$, $L_1$, and $L_2$. Each row reports the average
        gating distribution over 50 randomly sampled examples from the
        corresponding VSI-Bench subtask. The upper five rows correspond to
        low-level perception tasks, while the lower three rows correspond to
        high-level reasoning tasks.
    }}
    \label{fig:per_task_gating}
\end{figure*}

\noindent\textbf{Low-Level Perception Tasks.}
As shown in the upper part of Fig.~\ref{fig:per_task_gating}, low-level
perception tasks generally assign greater weight to shallow geometric
features at the early injection position. At $L_0$, object size and absolute
distance assign $37.4\%$ and $36.2\%$ of their weights to shallow features,
respectively. These tasks depend strongly on local object boundaries,
fine-grained depth variations, and metric cues. Relative distance and relative
direction exhibit more balanced shallow--intermediate distributions, assigning
$34.9\%$ and $33.7\%$ to shallow features and $35.6\%$ and $36.4\%$ to
intermediate features, respectively.

Room-size estimation shows a relatively stronger dependence on deep
representations than the other low-level tasks. Its deep-feature weight rises
from $32.4\%$ at $L_0$ to $42.3\%$ at $L_2$, reflecting the need to aggregate
scene-scale geometric information despite being categorized as a perceptual
task. Nevertheless, averaged across all five low-level tasks, shallow features
receive $35.0\%$ of the gating weight at $L_0$, confirming their importance
for fine-grained spatial perception.

\noindent\textbf{High-Level Reasoning Tasks.}
The lower part of Fig.~\ref{fig:per_task_gating} reveals a progressively
stronger preference for deep geometric representations as reasoning
complexity increases. At $L_2$, object counting, appearance order, and route
planning assign $42.9\%$, $46.8\%$, and $50.7\%$ of their weights to deep
features, respectively. Object counting requires the model to aggregate
objects across views while maintaining identity consistency. Appearance-order
reasoning additionally requires temporal and cross-view scene integration,
leading to a stronger preference for global geometric representations. Route
planning exhibits the largest deep-feature weight because it requires
holistic scene understanding and multi-step reasoning over navigable spatial
structures.

The distinction is already visible at $L_0$. Object counting remains
relatively balanced, with weights of $30.8\%$, $35.7\%$, and $33.5\%$ for
shallow, intermediate, and deep representations, respectively. In contrast,
appearance order and route planning assign their largest $L_0$ weights to
deep features, reaching $36.9\%$ and $39.9\%$. This indicates that globally
structured geometric information is useful from the beginning of the
reasoning process for cognitively demanding tasks.

\begin{figure*}[!t]
    \centering
    \includegraphics[width=0.95\textwidth]
    {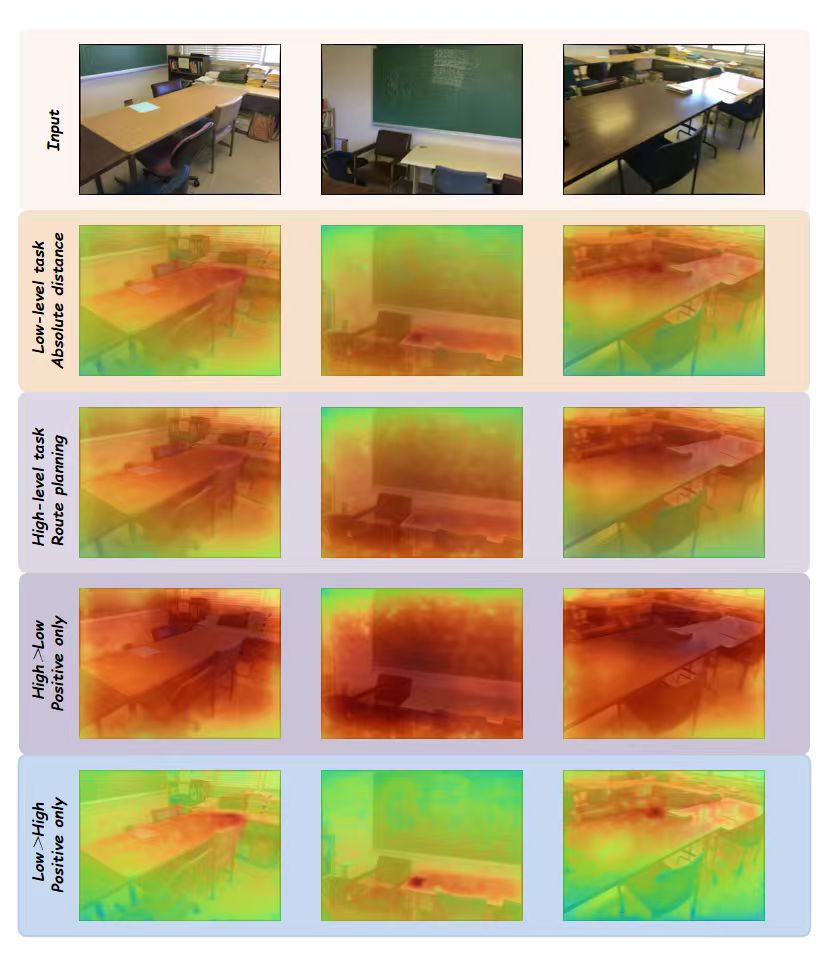}
    \caption{\small{
        \textbf{Task-Dependent Geometric Response Visualization on VSI-Bench.}
        The low-level absolute-distance task emphasizes fine-grained local
        geometry, whereas the high-level route-planning task produces broader
        responses over global scene structures.
    }}
    \label{fig:task_attention}
\end{figure*}

\noindent\textbf{Layer-Wise Selection Pattern.}
As shown in Fig.~\ref{fig:per_task_gating}, GeoFus exhibits a consistent local-to-global selection pattern across the three LLM injection positions. From $L_0$ to $L_2$, the weight assigned to shallow geometric features decreases across all eight tasks, whereas the contribution of deep features progressively increases. Intermediate features remain comparatively stable, receiving approximately $33\%$--$43\%$ of the total weight across tasks and injection positions. This pattern suggests that intermediate representations provide a persistent bridge between fine-grained geometric details and global scene structure.

The gating distributions also vary systematically across task groups. At $L_0$, low-level perception tasks assign an average weight of $35.0\%$ to shallow features, compared with $28.8\%$ for high-level reasoning tasks. Conversely, at $L_2$, high-level tasks allocate $46.8\%$ of their weights to deep features, whereas low-level tasks allocate $39.3\%$. These results indicate that GeoFus does not employ a fixed geometric fusion pattern. Instead, it adjusts the relative contributions of shallow, intermediate, and deep representations according to both the injection position and the spatial requirements of each task.

\subsection{Task-Dependent Geometric Response Visualization}
To further investigate how GeoFus adapts to different spatial reasoning requirements, we visualize its attention responses for a low-level absolute distance task and a high-level route planning task. As shown in Fig.~\ref{fig:task_attention}, the second and third rows present the attention maps for the two tasks, respectively. The fourth row retains only regions receiving stronger attention in the high-level task than in the low-level task, whereas the fifth row shows the opposite comparison. Route planning places greater emphasis on the global scene layout, navigable regions, and spatial relations among objects. In contrast, absolute distance estimation focuses more on local object geometry, boundaries, and nearby reference cues. These complementary attention patterns demonstrate that GeoFus adjusts its geometric focus according to task-specific reasoning requirements, further supporting its effectiveness in task-adaptive geometric fusion.

\subsection{Qualitative Results across VSI-Bench Tasks}

\begin{figure*}[!p]
    \centering

    \vsiresult{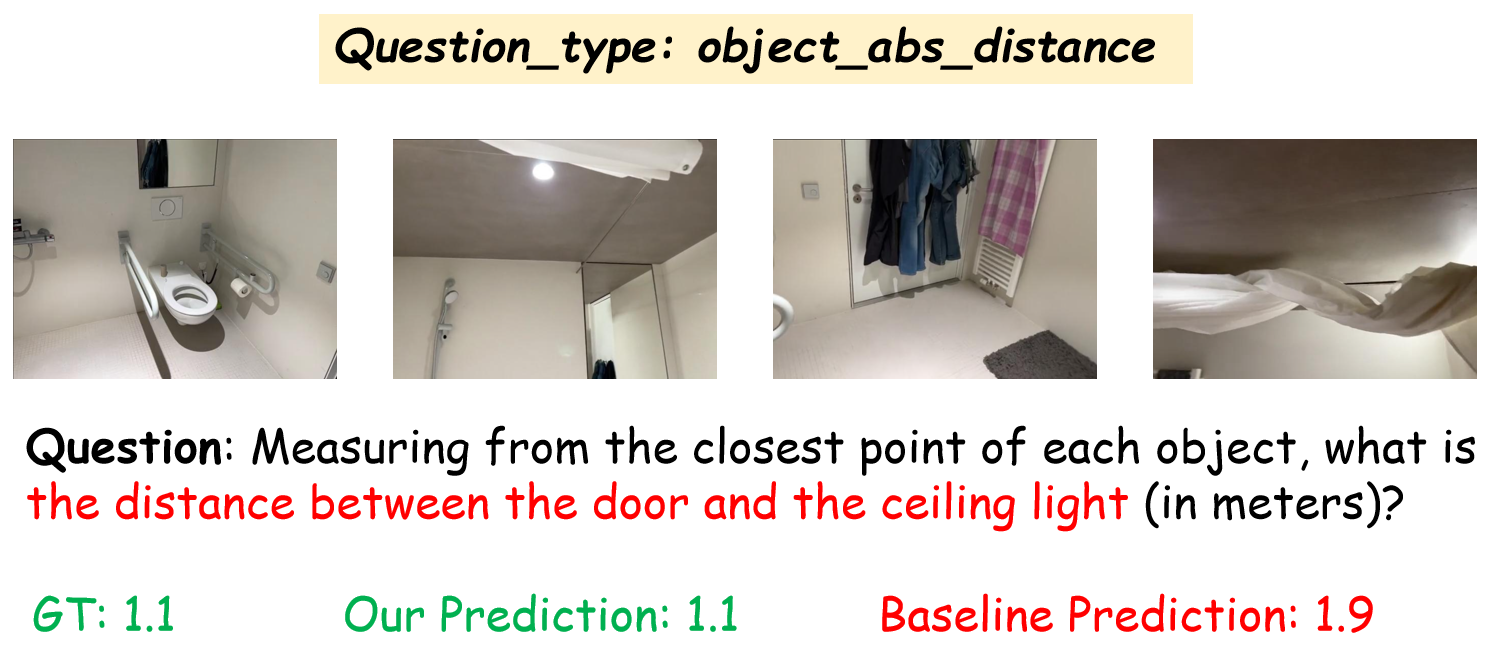}

    \vspace{2pt}

    \vsiresult{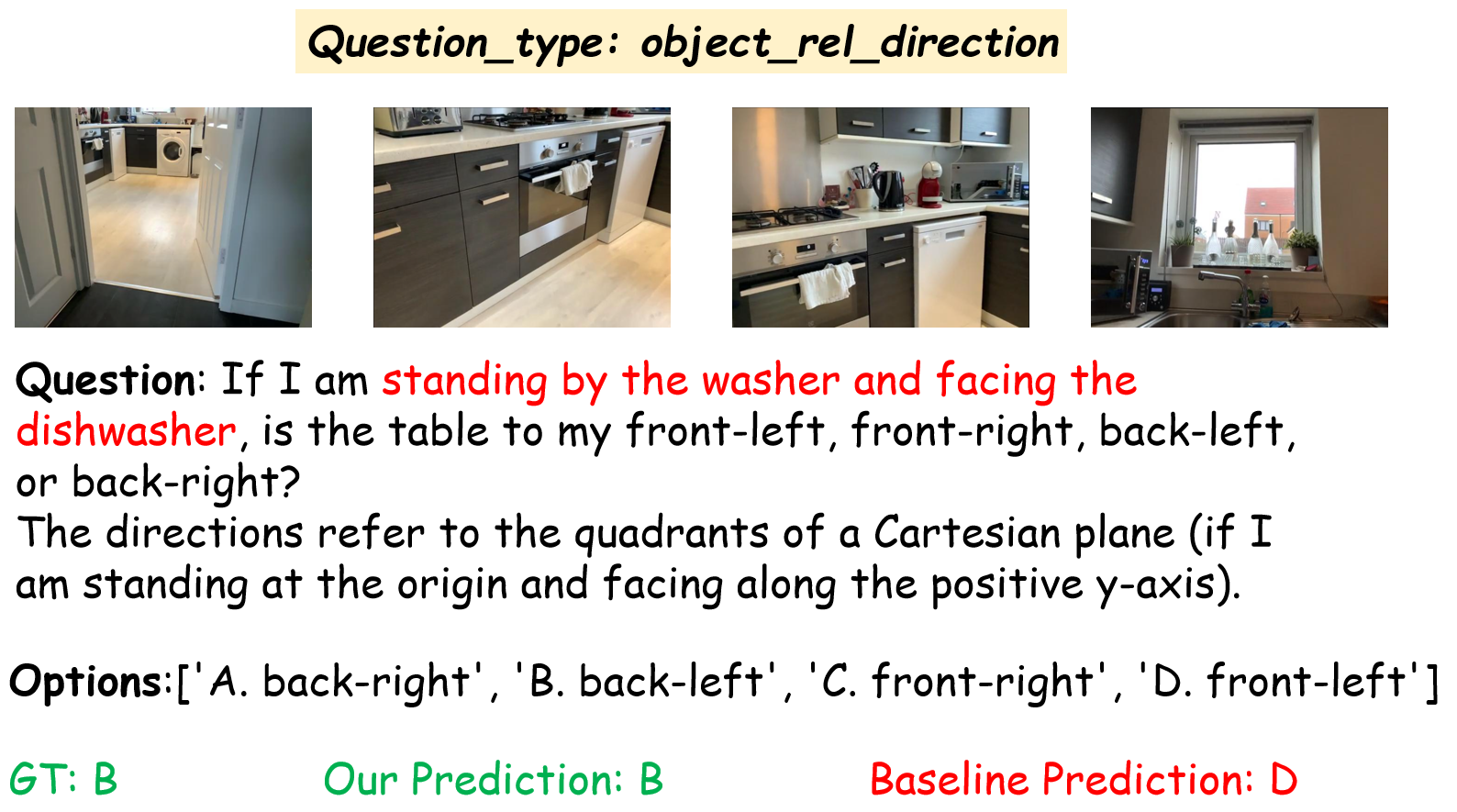}

    \vspace{2pt}

    \vsiresult{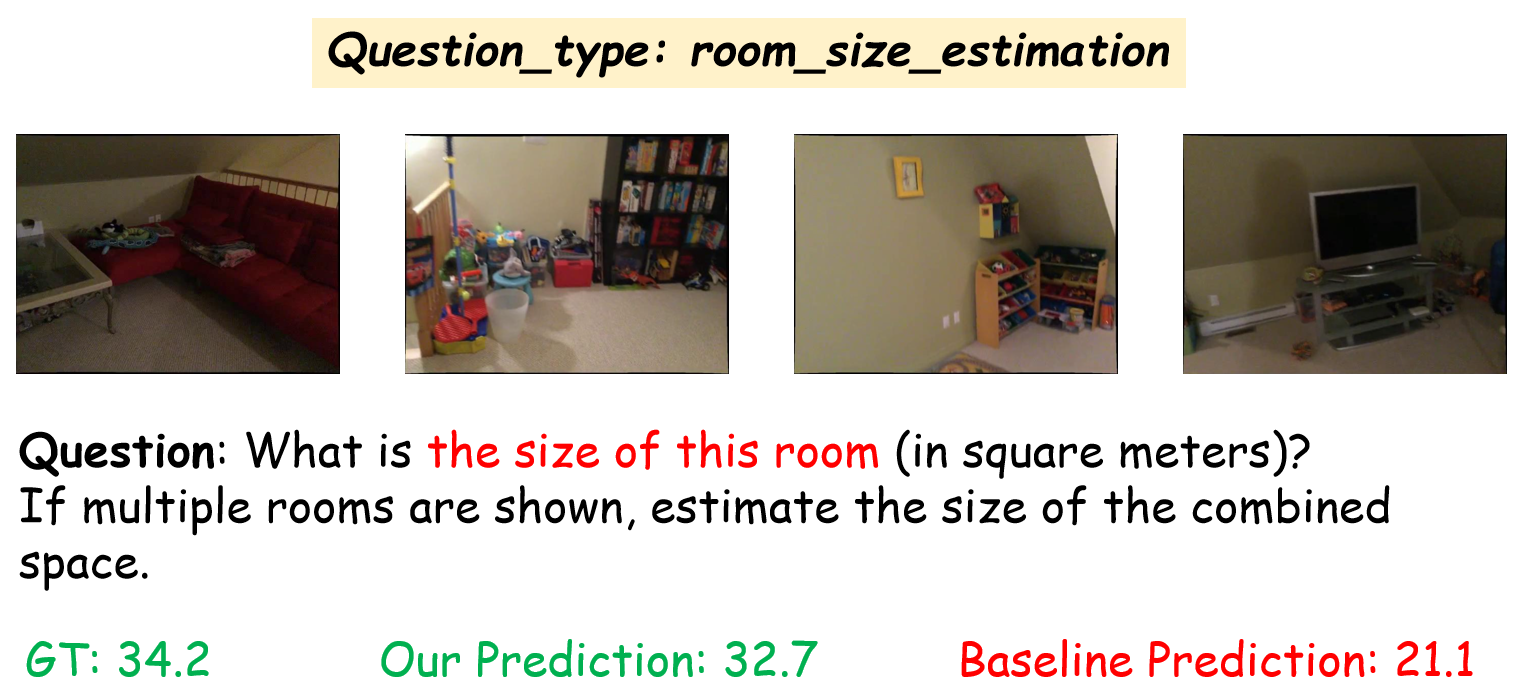}

    \caption{\small{
        \textbf{Qualitative Results on Low-Level VSI-Bench Tasks.}
        Representative comparisons on absolute-distance estimation,
        relative-direction reasoning, and room-size estimation. SpatialMind
        produces predictions consistent with or closer to the ground truth,
        whereas the baseline exhibits substantial metric and directional
        errors.
    }}
    \label{fig:vsi_low_level_qualitative}
\end{figure*}

\begin{figure*}[!p]
    \centering

    \vsiresult{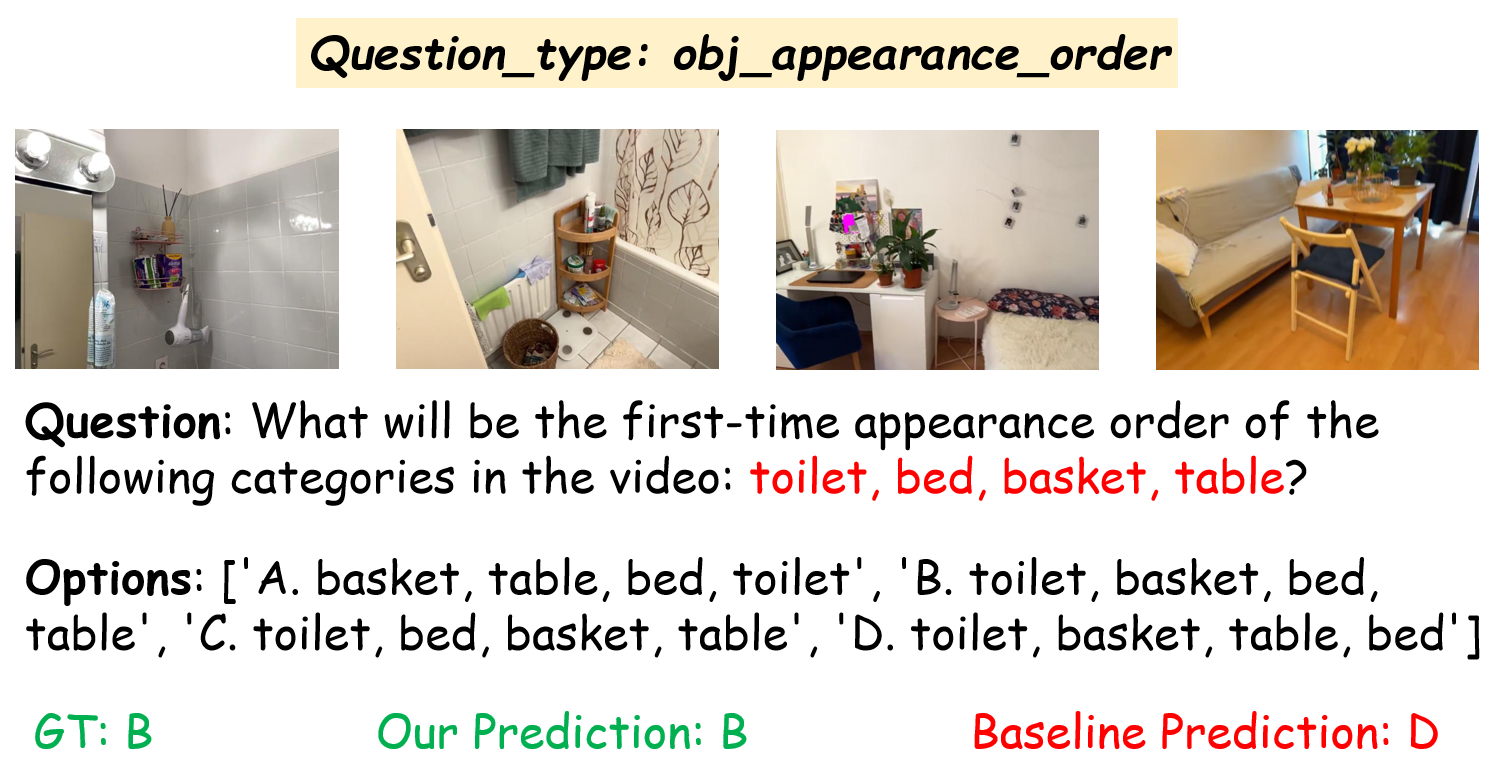}

    \vspace{2pt}

    \vsiresult{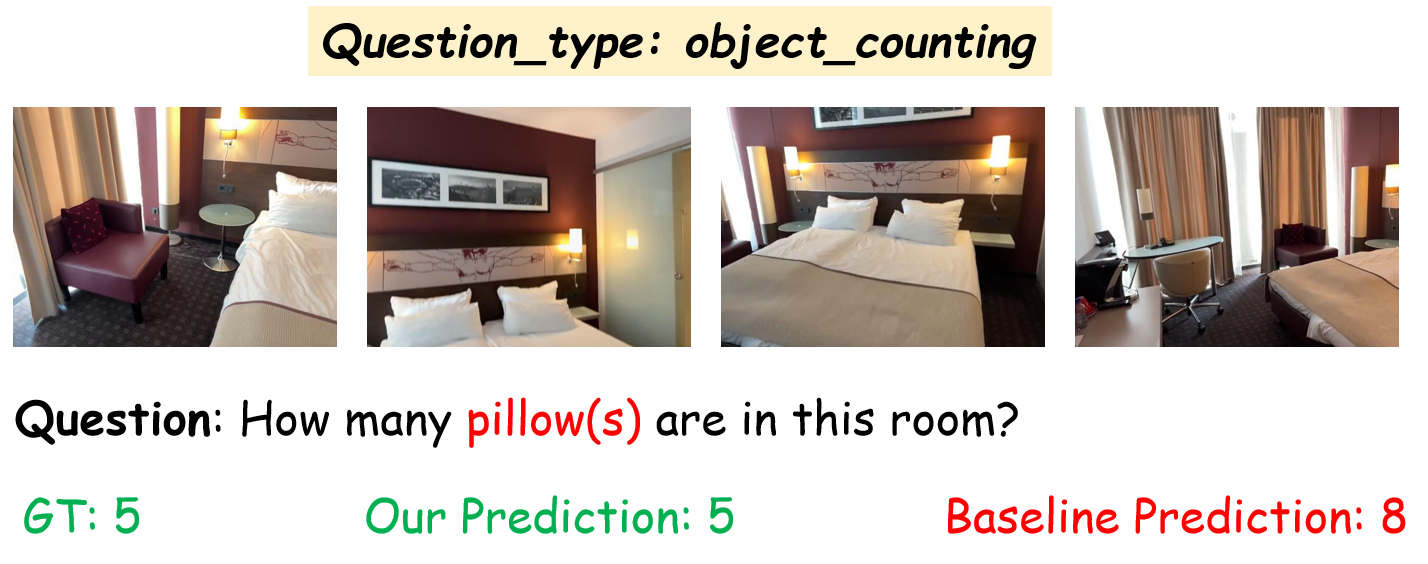}

    \vspace{2pt}

    \vsiresult{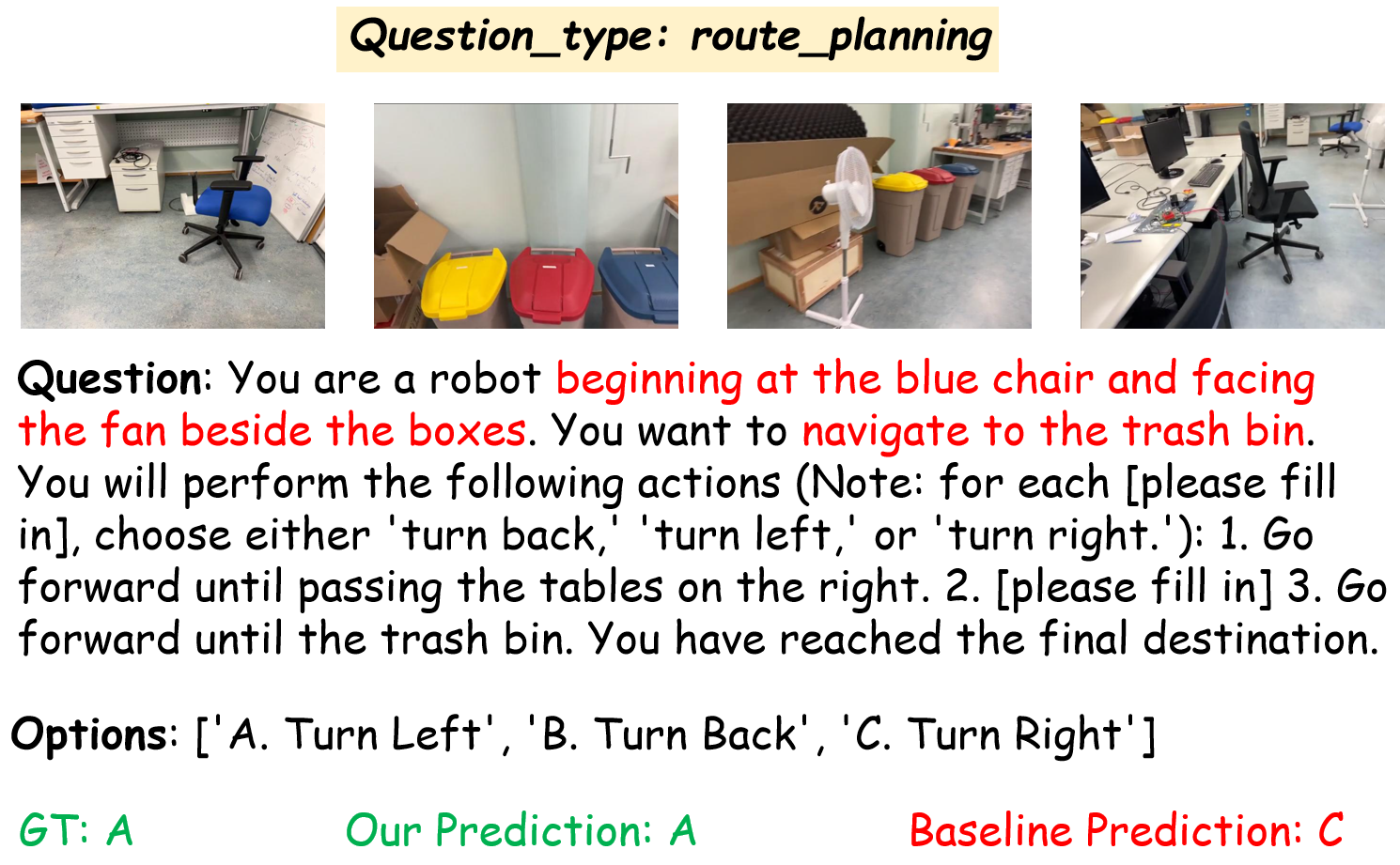}

    \caption{\small{
        \textbf{Qualitative Results on High-Level VSI-Bench Tasks.}
        Representative comparisons on appearance-order reasoning,
        cross-view object counting, and route planning. SpatialMind correctly
        integrates information across views and performs global or multi-step
        spatial reasoning, while the baseline produces incorrect predictions.
    }}
    \label{fig:vsi_high_level_qualitative}
\end{figure*}

Figures~\ref{fig:vsi_low_level_qualitative}
and~\ref{fig:vsi_high_level_qualitative} present six representative
comparisons on VSI-Bench. Figure~\ref{fig:vsi_low_level_qualitative} shows
three low-level perception tasks: absolute-distance estimation,
relative-direction reasoning, and room-size estimation. SpatialMind accurately
estimates the distance between the door and the ceiling light, correctly
determines the relative direction of the table, and predicts a room size close
to the ground truth, whereas the baseline produces substantial metric and
directional errors. Figure~\ref{fig:vsi_high_level_qualitative} presents three
high-level reasoning tasks: appearance-order reasoning, cross-view object
counting, and route planning. SpatialMind correctly integrates information
across frames to recover the appearance order, avoids duplicate counting
across views, and selects the correct navigation action, while the baseline
fails in all three cases. These qualitative results demonstrate that GeoFus
effectively adapts the granularity of geometric representations to different
task requirements, supporting both fine-grained spatial perception and global,
multi-step spatial reasoning.

\end{document}